\documentclass[lettersize,journal]{IEEEtran}
\usepackage{amsmath,amsfonts}
\usepackage{algorithm}
\usepackage{array}
\usepackage[caption=false,font=footnotesize,labelfont=sf,textfont=sf]{subfig}
\usepackage{textcomp}
\usepackage{stfloats}
\usepackage{url}
\usepackage{verbatim}
\usepackage{graphicx}
\usepackage{cite}

\usepackage{booktabs}
\usepackage{multirow}
\usepackage{tabularx}

\usepackage{hyperref}

\usepackage{algpseudocode}
\usepackage{amsmath}
\usepackage{color}
\usepackage{makecell}

\hypersetup{
colorlinks=true,
citecolor=black,
linkcolor=black
}

\begin{document}

%\title{SCMNet: A Scale-Complementary Multi-task Network for Object Detection in Aerial images}
\title{SCLNet: A Scale-Robust Complementary Learning Network for Object Detection in UAV Images}

\author{
            Xuexue Li,
    	    Wenhui Diao,
		    Yongqiang Mao,
            Xinming Li,
            and Xian Sun,~\IEEEmembership{Senior Member,~IEEE}

\thanks{This work was supported by the National Key R\&D Program of China under 2022ZD0118402.
\textit{ (Corresponding author: Wenhui Diao.) }
}

\thanks{X. Li, W. Diao, Y. Mao, X. Li and X.Sun are with the Aerospace Information Research Institute, Chinese Academy of Sciences, Beijing 100190, China, also with the Key Laboratory of Network Information System Technology (NIST), Aerospace Information Research Institute, Chinese Academy of Sciences, Beijing 100190, China, also with the University of Chinese Academy of Sciences, Beijing 100190, China, and also with the School of Electronic, Electrical and Communication Engineering, University of Chinese Academy of Sciences, Beijing 100190, China (e-mail: lixuexue20@mails.ucas.ac.cn; diaowh@aircas.ac.cn}}

%\thanks{Manuscript received April 19, 2021; revised August 16, 2021.}}



\maketitle

\begin{abstract} %Scale challenge including scale variation and small objects plague in object detection in UAV  images.
%Most of the recent UAV (Unmanned Aerial Vehicle) detector focuses more on general challenges such as uneven distribution and occlusion, while the neglect of the scale challenge that encompasses scale variation and small objects has led to its still plaguing object detection in UAV images.
Most recent UAV (Unmanned Aerial Vehicle) detectors focus primarily on general challenge such as uneven distribution and occlusion. However, the neglect of scale challenges, which encompass scale variation and small objects, continues to hinder object detection in UAV images.
Although existing works propose solutions, they are implicitly modeled and have redundant steps, so detection performance remains limited. And one specific work addressing the above scale challenges can help improve the performance of UAV image detectors. Compared to natural scenes, scale challenges in UAV images happen with problems of limited perception in comprehensive scales and poor robustness to small objects. We found that complementary learning is beneficial for the detection model to address the scale challenges.  Therefore, the paper introduces it to form our scale-robust complementary learning network (SCLNet) in conjunction with the object detection model. The SCLNet consists of two implementations and a cooperation method. 
In detail, one implementation is based on our proposed scale-complementary decoder and scale-complementary loss function to explicitly extract complementary information as complement, named comprehensive-scale complementary learning (CSCL). Another implementation is based on our proposed contrastive complement network and contrastive complement loss function to explicitly guide the learning of small objects with the rich texture detail information of the large objects, named inter-scale contrastive complementary learning (ICCL). In addition, an end-to-end cooperation (ECoop) between two implementations and with the detection model is proposed to exploit each potential. In short, SCLNet forms a more comprehensive representation through feature complementary and improves the representation of small objects through inter-scale contrast, which in turn comes to improve scale robustness and detection performance. Thorough experiments prove the effectiveness of our SCLNet on Visdrone and UAVDT datasets, including the fact that the novel components included in SCLNet are effective and competitive with many CNN-based and transformer-based methods, among other aspects. In general, our SCLNet can effectively address scale challenges and is a competitive model in UAV image object detection.
\end{abstract}


\section{Introduction}
\IEEEPARstart{O}{bject} detection is an important research task in computer vision, and in particular, is the cornerstone of many other tasks, such as behavior recognition\cite{hu2019driving}, instance segmentation\cite{chen20223}, visual object tracking\cite{zhao2022satsot}. Thanks to deep learning, object detection in natural scenes \cite{lin2014microsoft} achieves superior performance in recent years. However, for object detection of UAV (Unmanned Aerial Vehicle) images, generic detectors\cite{cai2018cascade,han2021align,ma2022feature} are difficult to migrate because of some problems such as occlusion, scale variation, and small objects\cite{li2021detection}.
The scale challenges including problems of scale variation and small objects, etc., are the well-known challenge that still plagues UAV image object detection. Although recent works\cite{li2023ogmn,li2020density,deng2020global,yu2021towards,zhang2019fully,duan2021coarse,yang2019clustered,yang2022querydet,du2023adaptive} have observed this challenge and proposed some solutions, these solutions are implicitly modeled, cumbersome in terms of detection process and therefore detection performance is still limited. Such limitations are detrimental to practical applications, and in the case of real-world UAV operations, a UAV detection model that accounts for all scales in the UAV's varying tilted viewpoints is at the heart of a good experience of human-computer interaction, and thus a method for effectively locating scale challenges is urgently needed.

%GLSAN\cite{deng2020global} proposes the global-loca detetion process to address variable distribution problem, which first performs global detection, then clusters and crops sub-regions based on the results of global detection, then performs fine local detection, and finally fuses the results of global and local detection results. QueryDet\cite{yang2022querydet} proposes a cascade query mechanism to address small objects problem, which first predicts the coarse locations of small objects and then computes the accurate detection results using high-resolution features base on those coarse locations. Even though these works try to address the scale challenge, their proposed solutions are cumbersome in terms of detection process and still limited in detection performance.
\begin{figure}[t]
\centering
\subfloat[]{
\label{dataset_scale_show}
\includegraphics[scale=0.42]{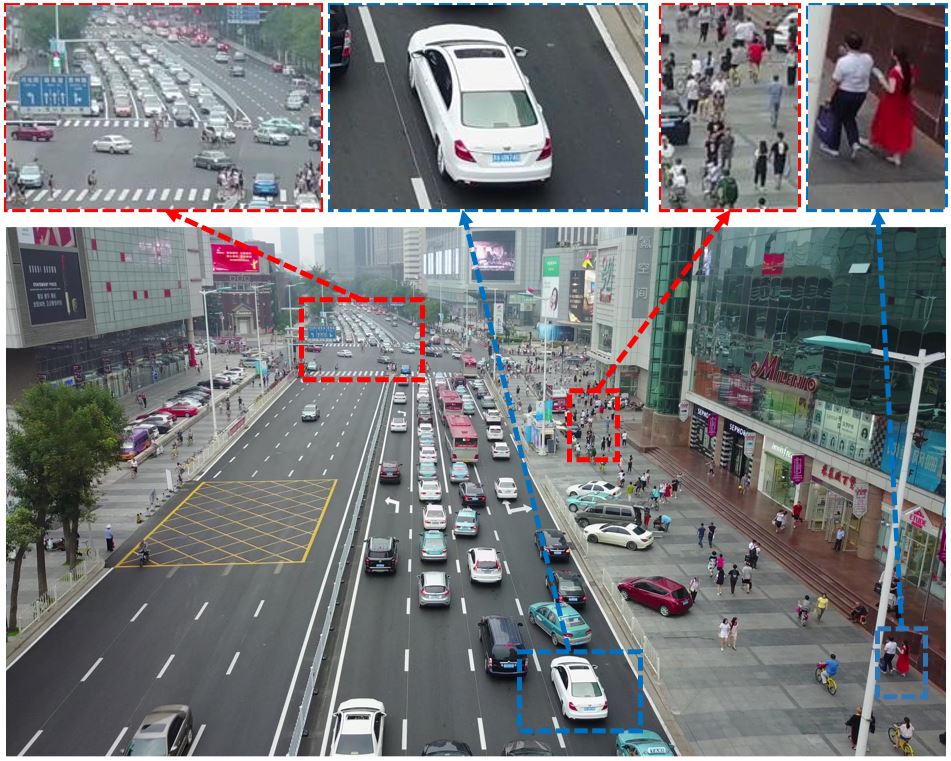}} \\
\subfloat[]{
\label{dataset_sta}
\includegraphics[scale=0.40]{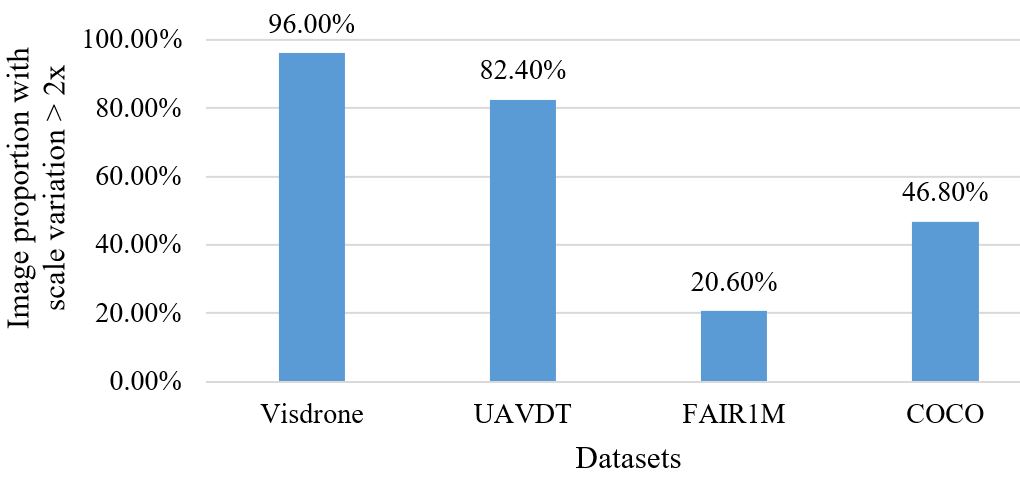}}
\caption{(a) Scale variation in UAV image. There is a huge difference in scale between larger and smaller objects of the same category of objects in the UAV image. (b) Comparison of image proportion with scale variation  $>$ 2x in different scene datasets. The image proportion with scale variation is much higher in the UAV scene than in other scenes.}
\label{fig_challenge}
\end{figure}
\begin{figure}[t]
\centering
\includegraphics[width=3.4in]{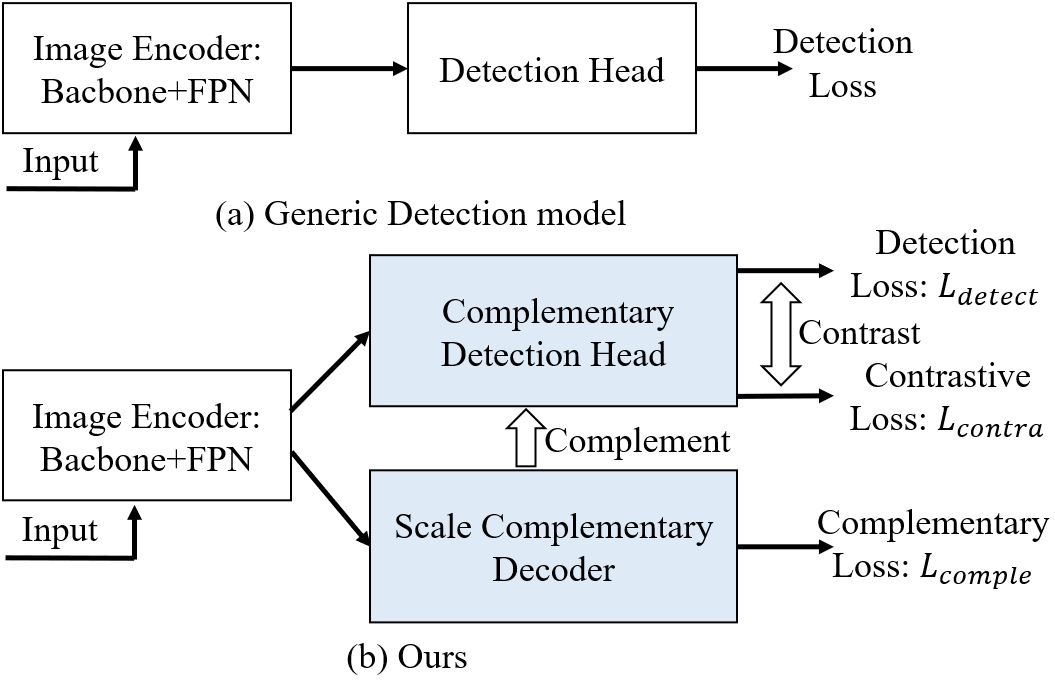}
\caption{Comparison of detection paradigms. (a) The generic detection paradigm. The detection head decodes the representation features extracted by the image encoder to output detection results, the detection performance of which is determined by the quality of the representation features. (b) Our proposed SCLNet. Complementary learning is introduced as a complement to form a comprehensive and robust representation.}
\label{fig:paradigm}
\end{figure}

The characteristics of the scale challenges in UAV images as a whole are generalised below: 
(1) \textbf{Limited perception in comprehensive scales.} The statistics in Figure\autoref{dataset_sta} clearly show that scale variation of objects in UAV images is more significant. Within the same category, instance objects vary in scale several times more than natural scenes\cite{lin2014microsoft}. The significant scale variation inevitably results in uneven number distribution in the wide range of scales. However existing detection models based on deep learning models tend to be more inclined to adapt to objects between a larger number of distributions\cite{cai2022novel}. This means that existing models are hard to represent and perceive all objects in comprehensive scales. We summarise this problem as the limited model perception in all scales under the significant scale variation. 
(2) \textbf{Poor robustness to small objects.}  There is a unique characteristic in UAV image object detection, while the small objects problem is a public problem in the computer vision recognition tasks\cite{cheng2023towards}. Both small objects and large objects often coexist in one UAV image due to scale variation, and the scale differences can be huge, as shown in Figure\autoref{dataset_scale_show}. The smaller objects lack textural information due to limitations such as low resolution, which makes the detection models hard to represent, but the opposite is true for larger objects. This characteristic is summarised as the poor robustness of the model to small objects. But a potential to be exploited is that the texture detail information is similar at different scales in intra-category objects. 
Due to these factors, existing models lack robustness to scale challenges, and the detection performance in UAV images still be limited. The significance of addressing the above issues is significant, and an effective approach can improve recall and reduce false alarms, which is very important for practical applications. Therefore, we propose a complementary approach to the encoding-decoding paradigm of the existing models, constituting a new paradigm to address the above problem, as shown in \autoref{fig:paradigm}.

%Image proportion with scale variation $>$ 2x in different scene datasets.
According to the characteristics of the scale challenges above, a novel scale-robust complementary learning network (SCLNet) is proposed to achieve better and more robust detection performance in UAV images.
%Based on the two problems in the analysis above, we propose a scale-complementary paradigm for object detection in UAV images to achieve better and more robust detection performance, named scale-robust complementary learning network(SCLNet). 
The generic detection models represent objects in an image in such a way that semantic features are first extracted using an image encoder, and then the detection head decodes the objects bounding boxes and categories based on the extracted semantic feature representations, as shown in \autoref{fig:paradigm}. The problem with this for UAV image object detection is that the complex scale challenges make it difficult for generic detection models to represent objects at all scales, which results in the characteristics mentioned above not being addressed. The paper found complementary learning to be beneficial in bridging these problems. 
The core of complementary learning in the human brain\cite{o2014complementary} is adding new information to existing information and constructing correlations to form memories, and similar deep learning work\cite{hou2020temporal} constructs complementary frame-to-frame learning in video to compensate for the lack of information in a single frame. 
Inspired by this paradigm, complementary learning is first introduced in UAV image detection, which includes two complementary learning implementations. 

Since the semantic features extracted by the original image encoder cannot satisfy the comprehensive scale representation of the objects, it is complemented with additional complementary information to form a comprehensive scale representation, which is our first implementation for solving limited perception in comprehensive scales, named comprehensive-scale complementary learning. Another implementation is to solve the problem of poor robustness to small objects. Objects of the same category in different scales in the same category have similar texture details. Since the model can perceive the rich texture details of large objects in the image, then use the knowledge learned from intra-category larger objects is complemented to small objects, making the model's perception of small objects more robust. Briefly, this paper introduces the advantages of complementary learning in UAV image object detection field by utilising the idea of complementary learning, through the lack of semantics to complement the existing semantics and the large scale to complement the small scale, and then form a robust scale representation, which ultimately improves the detection performance.

It is worth mentioning that this complementary learning approach we propose for boosting representations is explicitly modelled, unlike the implicit modelling included in most of the models already available. The meaning of implicit modelling here is relative to explicit modelling, implicit modelling focuses on improving the network structure, but compared to explicit modelling, the output of implicit modelling lacks explicit meaning, this paper, compared to the previous implicit modelling, improves the network structure at the same time, adds explicit constraints during training phase, so that the output has a clearer meaning, and realizes explicit modelling. In essence, our SCLNet consists of three components:

Firstly, in order to extract the complementary semantic information, we design a scale-complementary decoder that includes multi-scale perception with multiple convolutional kernels and multi-scale self-attention fusion for predicting scale complementary semantic features, and at the same time, we propose a scale-complementary loss function that constrains the area of the region where each instance is located to explicitly constrain the outputs and training of the decoder with the existing ground-truth. The essence of this implementation lies in our designed loss function using a priori information about object position and scale in the ground-truth to explicitly supervise the learning of the scale complementary decoder with strong representational capabilities during training. 
Secondly, in order for the model to transfer the knowledge learned from large objects to the learning of small objects, we design a contrastive complement network that includes category ground truth guided selection and feature interactions between small and large objects to combine the guidance of category ground truth to achieve the transfer, and we propose a contrastive complement lossfunction that constrains the consistency of the features output from the contrastive complement network branch and the original classification branch to transfer the knowledge learned from this network to the simple classification branch, making the training and inference consistent.
Finally, these two implementations mentioned above are independent of each other, and in order to exploit their potential even more and achieve better detection performance, we embed them into an existing detection model and design an end-to-end cooperation method (ECoop). Taken together with the specific implementations of the above components, our innovation in introducing complementary learning lies in compensating for the lack of representation of the scale challenges in a generic detector.

Thorough experiments conducted on Visdrone and UAVDT datasets prove the effectiveness of our SCLNet. The main advantages over existing methods are the following. Firstly, it is the enhancement of robustness to scale, both the ablation experiments and the visualisation of the detection results demonstrate that the performance of our proposed SCLNet for the detection of different scale objects is significantly improved. Secondly, it is the competitiveness, the comparison experiment proves that our proposed method is a competitive UAV image object detection model compared to both the CNN-based model and the transformer-based model. Finally, there is an improvement in the representation capability, and the visualisation of the feature maps demonstrates that our proposed method is not only capable of proposing a representation of the foreground objects, but also suppressing the background noise. 
The main contributions of this paper can be summarized as follows:

\begin{enumerate}
\itemsep=0pt
\item{We systematically analyse the characteristics of scale challenges in object detection in UAV images, and for the first time introduce complementary learning to model these problems.}
\item{We propose a comprehensive-scale complementary learning implementation for the problem of limited perception in comprehensive scales.}
\item{We propose an inter-scale contrastive complementary learning implementation for the problem of poor robustness to small objects.}
\item{We conduct thorough evaluation experiments on Visdrone and UAVDT datasets. It proves that our approach is not only effective in addressing scale challenge but also a competitive approach for UAV image object detection.}
%\end{list}
\end{enumerate}

The paper is organised as follows. In Section II, we review and discuss related works on UAV image object detection and complementary learning. In Section III, we discuss scale-robust representation modelling as well as introduce and explain our proposed scale-robust complementary learning network (SCLNet). In Section IV, we construct thorough experiments to evaluate the effectiveness of our proposed approach. Finally, we summarise the whole work and future research directions.

\section{Related work}

This section provides an overview of existing approaches in object detection for UAV images and complementary learning works. In \autoref{sec:over_object}, the typical recent works on UAV image object detection are listed, and contributions and shortcomings are illustrated. In \autoref{sec:complement_learn}, the discovery of complementary learning mechanism and the continued exploration of development in deep learning are listed and summarised in terms of their strengths and potential for locating the problems of this paper.

\subsection{Object detection in UAV images}
\label{sec:over_object}
The object detection task in UAV images is still challenging research, with detection performance still limited due to scale variation, small objects, and other challenges. A number of research works \cite{li2023ogmn,li2020density,deng2020global,yu2021towards,zhang2019fully,duan2021coarse,yang2019clustered} start with these challenges to improve UAV images detection performance. OGMN\cite{li2023ogmn} explicitly models occlusion between target objects to address occlusion challenge, achieving significant detection performance improvements, but its weakness is that ignores the scale challenges. UFPMP-Det\cite{huang2022ufpmp} designs a unified foreground packing pipeline and a multi-proxy learning to address small objects and uneven distribution issues, with the advantage that the improved process enhances the detection performance of small objects. TPH-YOLOv5++\cite{zhao2023tph} uses a sparse local attention to capture asymmetric information between detection heads, which implements a transformer detection header with stronger representations to address scale variations and motion blur. DMNet\cite{li2020density} incorporates cropping into the end-to-end detection model by clustering density maps, the advantage is providing a novel paradigm for the object distribution of UAV images. After the DMNet, some works such as GLSAN\cite{deng2020global}, DSHNet\cite{yu2021towards}, DREN\cite{zhang2019fully}, CDMNet\cite{duan2021coarse} and ClusDet\cite{yang2019clustered} come from uneven distribution and scale challenges to modify the detection process using cropping strategy to improve the detection performance. The main contributions are reviewed below. CDMNet\cite{duan2021coarse} designs a density estimation module to estimate coarse-grained density maps and then cropping with a clustering algorithm. GLSAN \cite{deng2020global} proposes a global-local detection process, clustering and cropping based on the global detection results, and then finely detects the cropped local regions. DSHNet\cite{yu2021towards} utilises the long-tailed distribution to optimise the detection head. ClusDet\cite{yang2019clustered} predicts several clustered regions and then resizes to refine detection. DREN\cite{zhang2019fully} trains a sub-network to predict regions of small objects for cropping. The Cascaded Zoom\cite{meethal2023cascaded} establishes a cascade two-stage detection process based on density guidance. The advantage of their works is optimising a robust method for the uneven distribution of objects in UAV images, but the weakness is the insufficient consideration of scale variations and too complex process.
%Although these works achieve some improvement in detection performance, the modified detection process is too complex.  
%, resulting in the disadvantages of redundant steps and high time consumption, which is extremely unfriendly to practical applications.  
In addition, some works\cite{yang2022querydet,du2023adaptive} optimize UAV image object detection models from different representation structures. The contribution of QueryDet\cite{yang2022querydet} is that uses query-based modelling approaches for the object detection task in UAV images, which achieve better detection performance. But the weakness is such a gain mainly benefits from the powerful representational capabilities of the transformer structure\cite{carion2020end, zhu2020deformable}. The contribution of CEASC\cite{du2023adaptive} is that proposes a novel context-enhanced sparse convolution to address the challenges of scale variation and small objects, but its weakness is still an implicit modelling approach that does not model the essence of the challenges in UAV images. Therefore, explicit modelling is implemented in the method constructed in this paper by constraining the output during the training phase to compensate for the above shortcomings.

\subsection{Complementary learning}%and Contrastive learning
\label{sec:complement_learn}
Complementary learning is helpful for constructing robust representation models. The mechanisms of complementary learning \cite{mcclelland1995there,mcclelland2013incorporating,o2014complementary,kumaran2016learning} have been keeping explored, and the recent introduction of complementary learning to deep learning \cite{song2018complementary,xu2020generative,zhang2018adversarial,hou2020temporal} tasks shows its significant advantages. Mcclelland et al.\cite{mcclelland1995there} reveal for the first time the complementary learning system between the hippocampus and the neocortex in the human brain memory. 
Then Mcclelland et al.\cite{mcclelland2013incorporating} use simulations to reveal the rapid learning of new schema-consistent information by the neocortex in complementary learning systems theory. 
Randall et al.\cite{o2014complementary} review the research history of the complementary learning framework and prove the vital theoretical force in brain memory systems of the complementary learning framework.
Simply put, complementary learning is an important mechanism of the human brain's memory system. For machine learning, Kumaran et al.\cite{kumaran2016learning} update the complementary learning systems theory to intelligent agents and propose complementary learning based on neuroscience for designing a machine learning network.
The series of works above charify that the human brain is capable of complementary learning, and can be updated to artificial intelligence such as machine learning. 
Innovative network structure inspired by the mechanisms of the human brain is an important approach to the development of artificial intelligence. Since the complementary learning mechanism in the human brain\cite{o2014complementary} is revealed, many AI research works focus on complementary learning, including natural language processing, computer vision, and many other areas. 
Song et al.\cite{song2018complementary} introduce complementary learning to achieve complementarity between multiple models, the advantage is that it improves quality of embedding in a word embedding task for natural language processing. Xu et al.\cite{xu2020generative} propose a generative-discriminative complementary learning for estimating labelling in semi supervision. Adversarial complementary learning is proposed by ACoL\cite{zhang2018adversarial} to achieve integral localization of objects in weak supervision. The contribution of these two works is to provide the new semi-supervised and weak-supervised ideas.
TCLNet\cite{hou2020temporal} proposes temporal complementary learning in the video pedestrian re-identification task, which mines information about inter-frame differences to achieve complementary. The advantage is that the representation of video data is more complete, but the weakness is that it can not be apply to the representation of images.

Although the above works demonstrate the potential of complementary learning in different areas, as far as we know, there is a lack of work that exploits complementary learning to accomplish robust UAV image semantic feature representation, especially for the challenge of significant scale variation, where it is difficult for a general model to extract robust semantic feature representation for objects at all scales. In detail, previous works focus on complementarity between temporal sequences in videos, but complementarity between objects within the same image is still lacking, which is the gap in the existing works regarding the use of complementary learning for UAV image object detection, and designing such an effective mechanism is critical for both addressing scale challenges and improving detection performance.
Inspired by both, we combine complementary learning and contrastive learning to address scale challenges including scale variation, small objects in UAV image object detection to achieve better detection performance. 
On the one hand, complementary learning is exploited to address scale variation by complementing the features of all-scale objects thereby forming a representation enhancement at most scales. On the other hand, the complementary idea is exploited to compensate for the lack of representation of small objects with large objects of the same category, and thus to achieve consistency in training and inferencing using a contrastive learning approach. The above two components form the core part of the complementary learning paradigm proposed in this paper. In particular, both of the above methods are implemented as explicit approaches, i.e., during the training process, explicit constraints are given, which are designed based on the manual labelled ground truth, and the predictions of the network have a relatively well-defined meaning, which can be effectively augmented to target weak representational parts, thus forming a robust representation to improve the detection performance.
%Although much of the works described above attempt to address scale challenge including scale variation or small objects in the detection process or representation structure, they suffer from some shortcomings, such as redundant steps, high time complexity, and lack of explicit modelling persist. In this paper, an explicitly modeled scale challenge, efficient paradigm is proposed to improve detection performance in UAV images. 
\begin{figure*}[t]
\centering
\includegraphics[width=6.8in]{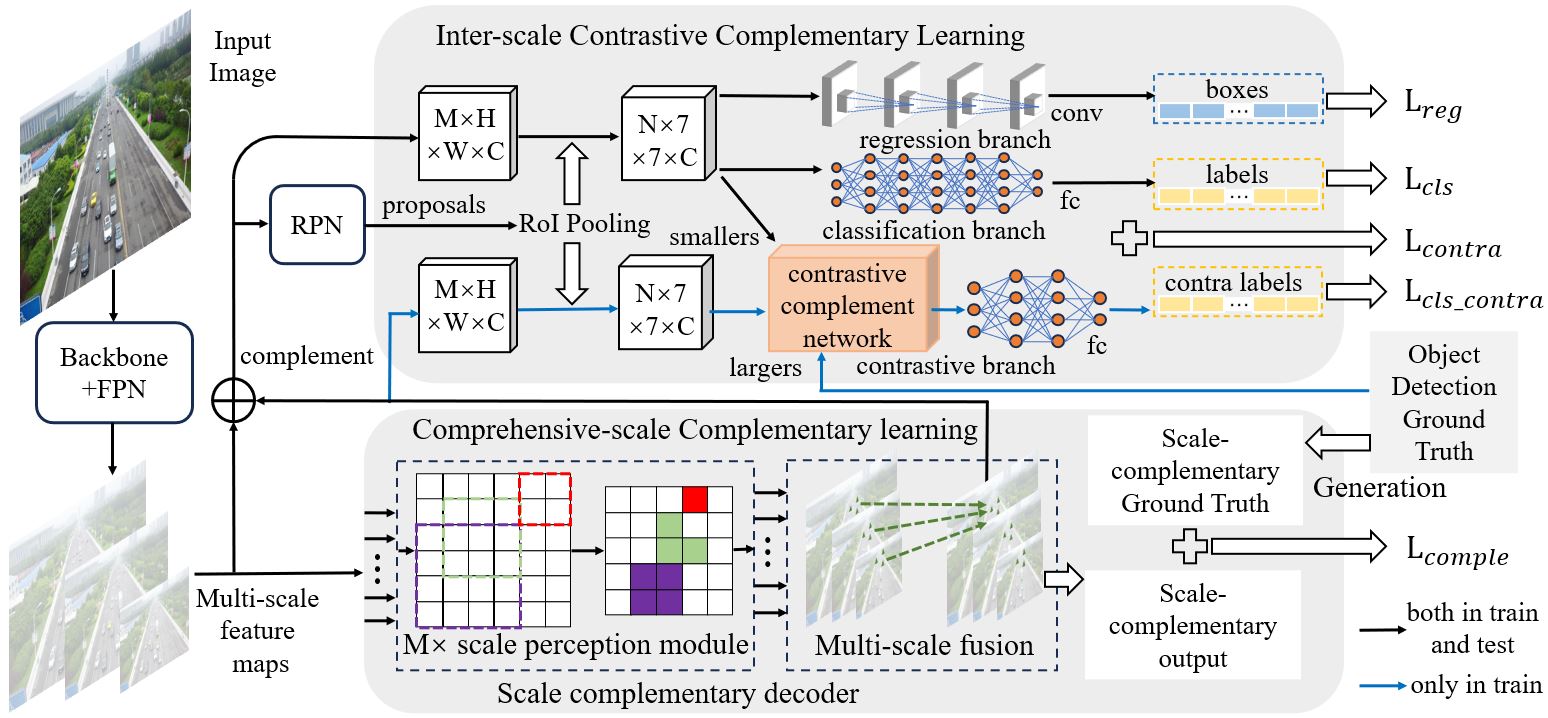}
\caption{
%Model structure of our proposed approach, which consists of two complementary learning implementations. One is comprehensive-scale complementary learning, with a scale complementary decoder and a scale-complementary loss function $L_{comple}$ implementation for extracting scale-complementary outputs as complements. The another is intra-category contrastive complementary learning, which is implemented by contrastive complement network and contrastive complement loss function $L_{contra}$, where the large objects in the same category is used to guide the learning of the small objects. $M$ denotes the number of multi-scale feature maps from image encoder (Backbone and FPN). $N$ denotes the number of proposals.
The framework of our SCLNet, consists of two main complementary learning implementations. One is comprehensive-scale complementary learning, which extracts scale-complementary outputs as complements. The another is intra-category contrastive complementary learning, which utilizes the large objects to guide the learning of the small objects. $M$ and $N$ denote the number of multi-scale feature maps and proposals respectively.}
\label{fig:model_all}
\end{figure*}
\section{Method}
It is as illustrated in \autoref{fig:model_all}, our proposed scale-robust complementary learning network (SCLNet) aims to implement a complementary learning paradigm with multi implementations for UAV image object detection to build a robust representation, which in turn improves detection performance, which consists of three key techniques: comprehensive-scale complementary learning (CSCL) as one implementation, inter-scale contrastive complementary learning (ICCL) as another implementation and end-to-end cooperation (ECoop). In \autoref{sec:representation}, a scale-robust UAV image object detection model is modelled. The CSCL is proposed to explicitly extract comprehensive semantic features of all scale objects for complementary use (\autoref{Complementary_learning}). The ICCL is proposed to be used to implement comparative complementary learning of large objects to small object instruction (\autoref{sec:Contrastive_learning}). The ECoop encompasses the cooperative methods of the above components and object detection model to achieve better detection performance (\autoref{sec:end-to-end}).

\subsection{Scale robust representation for UAV image object detection}
\label{sec:representation}
The scale challenges such as scale variation and small objects still plague the UAV image object detection. To better address the scale challenges, in this section, we analyse the characteristics of the scale challenges and construct a scale robust representation modelling with complementary learning for UAV image object detection. 

The scale variation in UAV images is intra-category and inter-category variation, and the scale of intra-category objects in the same UAV image can vary by more than a factor of ten. For larger scale objects, having a larger number of pixels in the input image and having a higher resolution mean that the texture details are more informative and it is easier for the model to extract discriminative features. While for smaller scale object, the fewer pixels and low resolution result in a lack of texture detail information, making it difficult for the model to extract discriminative features\cite{cheng2023towards}. This problem can be summarised as the model's lack of scale robustness. It is worth mentioning that although small scale is a common challenge in object detection in many research fields, such significant scale variation in UAV images is rarely found in other research fields. For the number of objects, the number distribution of the objects varies greatly in terms of scale, i.e., some scales have a larger proportion of objects and others have fewer, and for models based on deep feature learning they follow the tendency to fit the scales with a larger number distribution, especially for challenges with such significant scale variation in UAV images, and it is difficult for models to fit objects at all scales. This is another indication of the lack of scale robustness of existing UAV image detection models. Given input UAV image $I$, the problem can be described as:
\begin{equation}
\begin{aligned}
p(R|F_{enc}) &= \frac{p(F_{enc}|R)}{p(F_{enc})},& F_{enc}= E(I) \\
\end{aligned}
\end{equation}
where the semantic features extracted by the image encoder are defined as $F_{enc}$. And $E(\cdot)$ indicates the image encoder. According to bayes' law, the detection result can be characterised for $p(R|F_{enc})$ from $F_{enc}$. The problem is that for certain scales of objects, the extracted $F_{enc}$ is incomprehensive, and therefore the detection performance is limited.

For a scale-robust model for UAV image object detection, it should be able to perfectly address scale challenges including scale variation, small scales, etc. In other words, on the one hand, the model is able to robustly extract discriminative features for all scale objects, including smaller scale objects and larger scale objects in scale variation, and on the other hand, the model is able to fit objects at all scales where the number of samples is unevenly distributed. To achieve the purpose of scale robustness, we introduce the idea of complementary learning to achieve both. Such an idea can be described as:  
\begin{equation}
\begin{aligned}
F_{enc} = E(I),& F_{comple} = D_{comple}\big(E(I)\big)  \\
p(R|F_{enc}, F_{comple}) &= \frac{p(F_{comple}|R, F_{enc})\cdot p(R|F_{enc})}{p(F_{enc})}\\
\end{aligned}
\end{equation}
where the complementary semantic feature used to compensate for the incomplementary shortcomings of $F_{enc}$ is denoted as $F_{comple}$. And $p(F_{comple}|R, F_{enc})$ is the probability of extracting scale complementary semantic information given $F_{enc}$ and detection results. $p(R|F_{enc})$ is the probability of obtaining detection results given $F_{enc}$. The detection results are the position with the greatest posterior probability.
\begin{equation}
\begin{aligned}
R^{\ast} &= \mathop{\mathrm{argmax}}\limits_{R}\Big(p(R|F_{enc}, F_{comple})\Big)\\
         &= \mathop{\mathrm{argmax}}\limits_{R}\Big(p(F_{comple}|R, F_{enc})\cdot p(R|F_{enc})\Big)\\
\end{aligned}
\end{equation}
Thus, the detection performance is determined by the likelihood model $p(F_{comple}|R, F_{enc})$ and the spatial prior $p(R|F_{enc})$. The key components of SCLNet proposed in this paper is designed around this paradigm. The roadmap for the entire paradigm is as follows. The two main components are its core: firstly, an explicit complementary learning network $D_{comple}(\cdot)$ is designed to compensate for the fact that the perceptual capabilities of existing models are difficult to cover objects at all variable scales (detailed in \autoref{Complementary_learning}), which is explicitly modeled for the likelihood model $p(F_{comple}|R, F_{enc})$ and as one implementation of complementary learning; secondly, we use complementary learning to exploit larger objects in UAV images to supervise the model to extract robust representation for smaller objects (detailed in \autoref{sec:Contrastive_learning}), which is modeled for the spatial prior $p(R|F_{enc})$ and as another implementation. Finally, we combine the above proposed components with existing detection models to form an end-to-end detection model (detailed in \autoref{sec:end-to-end}). Overall, the first two components serve as the core of two complementary learning implementations, and the last one serves as collaboration and exploitation of the two mentioned above.

\subsection{Comprehensive-scale complementary learning}
\label{Complementary_learning}
In order to construct a comprehensive and robust feature representation for the UAV image object detection model, we construct a novel complementary learning in comprehensive scales based on the idea of complementary learning on the basis of the generic detection model for forming a comprehensive scale representation, named as comprehensive-scale complementary learning (CSCL). As a whole, the objective of the CSCL component is to extract explicit complementary features for all scales of objects in each category, to be used as a complement to form a more robust representation at all scales. The specific implementation of complementary learning in comprehensive scales we designed consists of two parts, one of which is a scale-complementary decoder for extracting scale-complementary semantic features. The other part is the scale-complementary loss function for supervised training the scale-complementary decoder.

\subsubsection{Scale-complementary decoder}
\begin{figure}[t]
\centering
\includegraphics[width=3.4in]{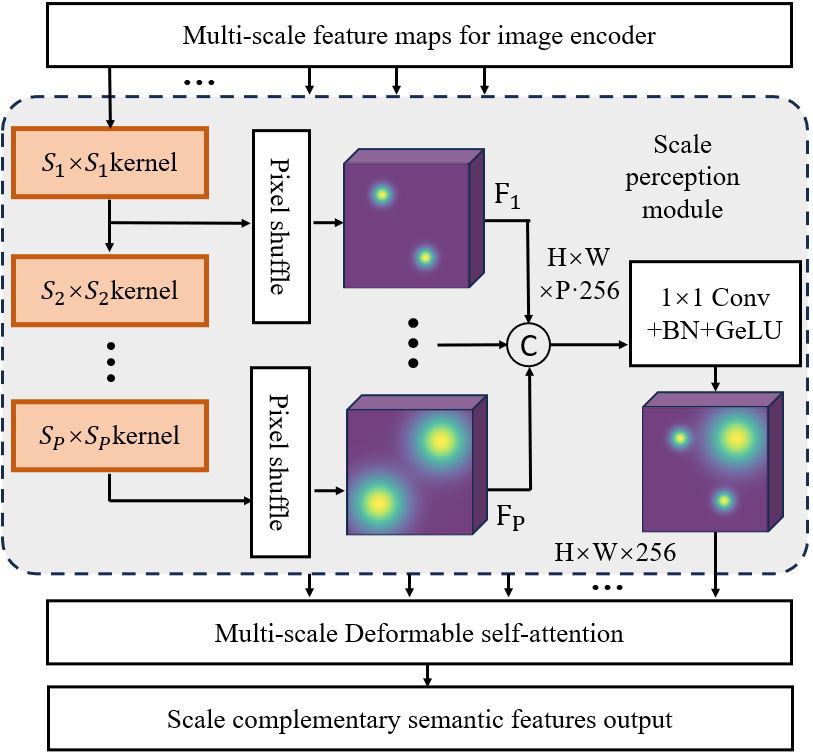}
\caption{The detailed structure of scale-complementary decoder. The multi-scale features of the image encoder are sequentially used as inputs. The convolutions with different scale kernels, pixel shuffle and multi-scale fusion are performed then. Finally, the multi-scale deformable self-attention is performed to get the final output.}
\label{fig:decoder}
\end{figure}
Existing detection models extract semantic features that are difficult to cover all scales of objects in UAV images, so we use the idea of complementary learning to build a comprehensive and robust feature representation for the full range of scales. The focus here is on extracting complementary semantic features for complementing existing semantic features, a module for extracting complementary semantic features is urgently needed, so we design the scale-complementary decoder for this purpose. The outputs of the scale-complementary decoder will be used as complements to form a comprehensive and robust representation, which is also relied upon by the other components of our proposed method, and thus the scale-complementary decoder is the key to comprehensive-scale complementary learning.

\begin{algorithm}[h]
\caption{The scale-complementary decoder:}
\label{algorithm_scalecomple}
\begin{algorithmic}[1]
\Require{multi-scale feature maps from image encoder $\{F_{enc}^0,\cdots,F_{enc}^{M}\}$}
\Ensure{scale complementary semantic feature maps $F_{comple}$}
\emph{Calculations of scale-complementary perception:}
\For {$F_{enc}^{i}$ in enumerate($F_{enc}$)}
\State$map_{i}^{0} \gets F_{enc}^{i}$
\For {$j$ in enumerate($p$)}
\State$map_{i}^{j} \gets Conv(map_{i}^{j-1}, kernel=S_{j})$
\State$map_{i}^{j} \gets PixelShuffle(map_{i}^{j})$
\EndFor
\State$F_{comple}^{i} \gets Concat(map_i^0,\cdots,map_i^p)$
\State$F_{comple}^{i} \gets ConvBNReLU(F_{comple}^{i})$
\EndFor\\
\emph{Interactions multiple scales:}
\State$F_{comple} \gets \{F_{comple}^{0},\cdots,F_{comple}^{M}\}$
\State$F_{comple} \gets MSDeformAttn(F_{comple})$\\
\Return $L_{comple}$
\end{algorithmic}
\end{algorithm}
The detailed design of the scale-complementary decoder is shown \autoref{fig:decoder} and Algorithm \autoref{algorithm_scalecomple}. The multi-scale feature maps output from the image encoder (Backbone and FPN) are sequentially adopted as inputs to the module sequentially at different scales, which has the advantage of not requiring an additional image encoder, which largely reduces the number of model parameters. Firstly, the decoder designs a scale perception module to achieve the perception of variable scale objects. The convolution of kernels of different sizes acting on the input feature maps are used to perceive objects of varying scales, and the size of these convolution kernels increases sequentially in order to achieve a progressive receptive field, determined based on the statistics of the object scales in the overall dataset, which allows for a more relevant perception. After obtaining multiple outputs from different convolutional kernels, we upsample to the same spatial dimensions using pixel shuffle\cite{shi2016real}, and then concatenate these outputs in channel dimensions before fusion-selective dimensionality reduction to obtain preliminary results. Secondly, we employ a multi-scale deformation self-attention module\cite{zhu2020deformable} to perform a multi-scale fusion of the preliminary results. The advantage of doing so is that the information between different feature maps can be interacted with because our proposed scale perception module is performed for each feature map, and such multi-scale interactive fusion can make the scale perception module work better for variable scale perception.

The scale-complementary decoder designed above is endowed with the potential to extract complementary semantic features across the comprehensive scales, and this is demonstrated in the experimental section.

\subsubsection{Scale-complementary loss function}
\label{sec:complementary_loss}
Although the above-mentioned scale-complementary decoder has the potential to extract complementary semantic features, if the model only relies on this module, the complementary learning achieved is still implicit modelling, which is similar to most existing implicit modelling methods\cite{wu2022casa}. The cause is because that the semantic information derived by the module does not present a clear interpretability. In order to explicitly model the scale challenges, that is, the comprehensive scale complementary learning we propose is highly interpretable, we propose a scale-complementary loss function for the supervised training of scale-complementary decoder. The performance of the other components is determined by the quality of the scale-complementary decoder output, so the scale-complementary loss function is critical.
\begin{algorithm}[h]
\caption{Calculation of the scale-complementary loss function:}
\label{algorithm1}
\begin{algorithmic}[1]
\Require{input image $I$; ground-boxes $Boxes$; scale complementary semantic feature outputs $F_{1}^{S},\cdots,F_{M}^{S}$; }
\Ensure{scale complementary loss $L_{comple}$.}\\
\emph{Initialization:}\\
\emph{$W,H \gets I.size.width, I.size.height$}\\
\emph{$GT_{scale} \gets Zero-matrix(W, H)$}\\
\emph{Generate scale complementary ground truth:}
\For {$box$ in enumerate($Boxes$)}
\State$map \gets Zeros(H, W)$
\State$map[box(x_{center}), box(y_{center})] \gets 1$
\State$sigma \gets \big(box(x_{center})*box(y_{center})\big)^{-2}$
\State$GT_{scale}+=GaussainBlur(map, sigma)$
\EndFor\\
\emph{Calculate the loss function:}
\For {$i, F_{i}^{S}$ in enumerate($F_{1}^{S},\cdots,F_{M}^{S}$)}
\State$h, w \gets F_{i}^{S}.size.height, F_{i}^{S}.size.width$
\State$GT_{scale} \gets DownSample\big(GT_{scale},(h, w)\big)$
\State$L_{i} \gets MSE(F_{i}^{S}, GT_{scale})$
\EndFor\\
\emph{Averaging loss values of multiple maps:}\\
\emph{$L_{comple} \gets \frac{1}{M}\sum_{i=1}^{M} L_{i}$}\\
\Return $L_{comple}  $
\end{algorithmic}
\end{algorithm}

The core idea of the scale-complementary loss function is to construct scale-complementary ground-truth, which is then used to guide the scale-complementary decoder to learn the desired predictive power. 
We utilize the object detection ground boxes to generate the scale complementary ground-truth with Gaussian blurring operations\cite{zhang2013gaussian} with different sized kernels around the centroid of the ground boxes in the UAV image, the size of the Gaussian kernel is proportional to the areas of the ground boxes, which corresponds to the convolution kernels of different sizes in the scale perception decoder. The scale-complementary loss function is calculated from the generated scale-complementary ground-truth and the outputs of the scale-complementary decoder. The detailed calculation of this loss function is described in Algorithm \autoref{algorithm1}.

A scale-complementary decoder with the potential to predict complementary semantic features at all scales and a scale-complementary loss function that implements explicit modelling constraints of scale challenges enable comprehensive-scale complementary learning. That is, the addition of the two components enables the model to predict complementary semantic features that are an important part of the comprehensive and robust representation of all scale objects. These two components are the core innovations of the CSCL component. Notably, after extracting the complementary information, we use a simple element-by-element summation to combine the extracted complementary information with the semantic information extracted by the existing image encoder as the basis for the detection head decoding. Thus, this simple element-by-element summation is different from the identity map in ResNet to combine and the objective is to utilise the complementary information.

\subsection{Inter-scale contrastive complementary learning}
\label{sec:Contrastive_learning}
In addition to the comprehensive scale complementary learning mentioned above, the paper also proposes a kind of complementary learning to compensate for small object representation. Briefly, the paper adopts the semantic information of large-scale objects to guide the model learning to optimise the perception of semantic information of small objects among the same category, so that the semantic information of small objects predicted by the model is more comprehensive and robust, namely inter-scale contrastive complementary learning (ICCL). Overall, the objective of the ICCL component is to utilise the rich semantics of the larger objects in each category to enrich the semantics of the smaller ones, resulting in a more robust representation of the smaller ones. The specific implementation contains two parts: network optimisation and loss function construction.

\subsubsection{Contrastive complement network}
In UAV images, large objects are rich in texture details, which are easier for the model to extract discriminative features, whereas for small objects, the lack of texture details makes it difficult for the model to extract comprehensive and robust discriminative features, but the texture details of the same category of objects are similar, and it is only due to the scale variation caused by the tilted pose of the UAV that such a difference is made. It is this difference that we propose to compensate for with our contrast-complemented component, and thus we designed the contrast-complemented network module. The purpose of our proposed contrastive complement component is precisely to compensate for this difference, and thus we design the network of contrastive complement for implementing inter-scale contrastive complementary learning.
\begin{figure}[t]
\centering
\includegraphics[width=3.2in]{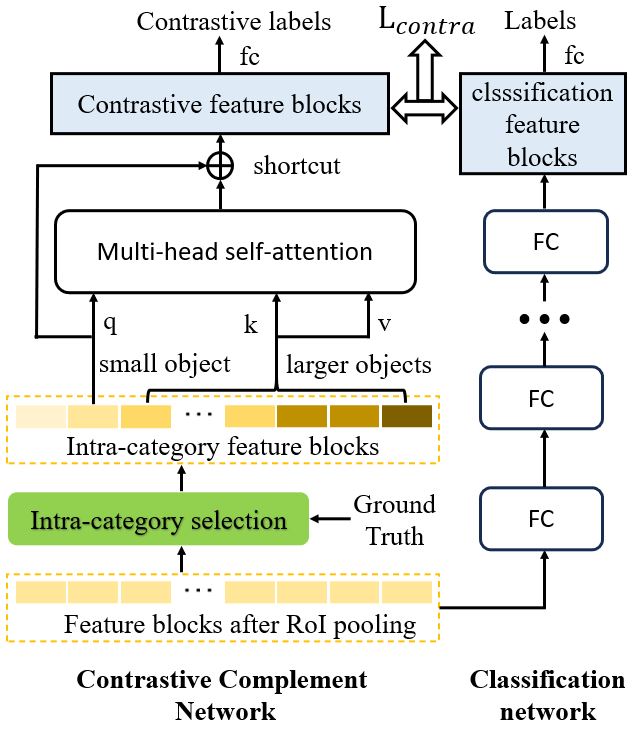}
\caption{
%The specifics of contrastive complement network. The contrastive complement network is parallel to the classification network in the detection head, and the feature blocks after roi pooling is assigned by category based on the ground truth, and the larger objects are used to complement the smaller object's feature blocks in the same category, outputting the contrastive feature blocks that are used to guide the classification network in learning about the small objects.
The specifics of the contrastive complement network, which is parallel to the classification network, and the feature blocks are assigned by ground category, and the larger objects complement the smaller object's feature blocks in the same category. The final output guides the classification network during the training phase.}
\label{fig:contrastive}
\end{figure}

\begin{algorithm}[h]
\caption{Contrastive complement network:}
\label{algorithm_contranet}
\begin{algorithmic}[1]
\Require{Feature blocks after RoI pooling $\{F_{enc}^0,\cdots,F_{enc}^{N}\}$; ground truth about category $GT_{category}$}; number of the category $K$
\Ensure{Contrastive feature blocks $\{F_{0}^{contra},\cdots,F_{N}^{contra}\}$}; contrastive labels $CLS_{contra}$\\
\emph{Intra-category selection:}
\For {$n, gt$ in enumerate($GT_{category}$)}
\State$k = gt.category number$
\State$set_{k}.append(F_{enc}^{n})$
\EndFor\\
\emph{Complements between intra-category instances:}
\For {$k$ in enumerate($K$)}
\State$ set_{cur} \gets sort(set_{k}, cmp=box.area)$
\For {$j, F_cur$ in enumerate($set_{cur}$)}
\State$q \gets F_cur; k, v  \gets set_{cur}[j+1:]$
\State$q \gets MultiHeadSelfAttn(q, k, v)$
\State$F_{cur} \gets shortcut(q, F_{cur})$
\EndFor
\EndFor
\State$\{F_{0}^{contra},\cdots,F_{N}^{contra}\} \gets Concat(set_{0},\cdots,set_{K})$\\
\emph{Predictions for contrastive labels:}
\State$CLS_{contra} \gets FC(F_{0}^{contra},\cdots,F_{N}^{contra})$
\State$F_{contra} \gets \{F_{0}^{contra},\cdots,F_{N}^{contra}\}$\\
\Return $F_{contra}, CLS_{contra}$
\end{algorithmic}
\end{algorithm}
As shown in \autoref{fig:contrastive} and Algorithm \autoref{algorithm_contranet}, the input to the network is the feature blocks after the roi pooling operation of all the proposals, this has the advantage of reducing the computation to a great extent, these proposals represent the spatial regions of the image where the target objects may be present, so the implementation with the pooled feature blocks possesses increases the interpretability of the model. Subsequently, we use the label assignment results to select the intra-category feature blocks corresponding to the objects within the same category. 
The label assignment here uses the existing max iou assigner method, which achieves the assignment of ground truth to proposals based on matching the maximum iou value of the proposal boxes and the ground boxes. The corresponding ground box that each proposal box gets assigned contains information about the category of ground truth, and our intra-category selection divides the proposals based on this category information of ground truth; the result of the division is the positive examples for each category, and the positive example proposals of the same category undergo a subsequent contrastive complement computation.
Such an intra-category selection makes the later perception guidance of the large objects to the small objects have the property of category decoupling, which is more conducive to fine-grained detection. With the intra-category feature blocks, one of the smaller objects is selected, and the larger objects are used as a reference to guide this smaller object, and these are fed into the multi-head self-attention module network\cite{dosovitskiy2020image} to enable complementary learning of smaller object in comparison to the larger objects. This is followed by a shortcut operation with the primitive small object feature blocks to extract the contrastive feature blocks.

\subsubsection{Contrastive complement loss function}
\label{sec:Contrastive_loss}
The inputs of the above contrastive complement network are involved with ground-truth inputs, but the ground-truth is not accessible in the inference phase, inspired by the knowledge distillation\cite{yang2022adaptive}, the paper proposes to distill the knowledge learned from the contrastive complementary network branch by using a lightweight classification network branch in the training phase, so the paper constructs a loss function to achieve this purpose, named as contrastive complement loss function. The contrastive complement loss function $L_{contra}$ can be formulated as:
\begin{equation}
\label{func:contra}
\begin{aligned}
L_{contra} = \frac{1}{N}\sum_{i=1}^{N}\left ( F_{i}^{contra}-F_{i}^{cls} \right )^{2}  \\
\end{aligned}
\end{equation}
where $F_{i}^{contra}$ and $F_{i}^{cls}$ indicate contrastive feature blocks and classification feature blocks respectively. $N$ denotes the number of the proposals.

By joining the collaboration of the contrastive complement network and the contrastive complement loss function, the model is relatively able to enhance the perceptual abilities of small objects, which compensates for the defects of the model in extracting discriminative features for small objects that are not comprehensive and robust, and thus improves detection performance.

\subsection{End-to-end Cooperation}
\label{sec:end-to-end}
The above two components comprehensive scale complementary learning(CSCL) and Intre-scale contrastive complementary learning(ICCL) are both key components of the proposed scale-complementary learning network in the paper, but the above constructed components are relatively independent in the model and do not collaborate sufficiently, and the potentials of the two components are not fully exploited. Therefore, based on the above two components CSCL and ICCL, this paper adds them to the existing detection model, and constructs a collaborative process of the two components, named end-to-end cooperation (ECoop). Overall, the objective of the ECoop component is to explore the more appropriate way of collaboration between the two components, CSCL and ICCL, for end-to-end target detection, in order to expect to achieve better object detection performance.

\subsubsection{Cooperation of CSCL and ICCL}
In order to better exploit the potential of the two components mentioned above to improve detection performance, we designed the cooperation methods between the two components, as shown in \autoref{fig:model_all}. 

On the one hand, we fuse the scale complementary semantic feature output of the scale complementary decoder with the multi-scale feature maps from the image encoder (Backbone+FPN), where the fusion is the most straightforward element-by-element summation. The advantage of direct element-by-element summation is that it does not require additional network involvement and does not introduce an increase in the number of model parameters. Since the extraction of the scale complementary semantic feature outputs is explicit and possesses an explicit interpretative essence, such a fusion can exploit the potential of the CSCL components. 
 
On the other hand, the original input to the ICCL network described in \autoref{sec:Contrastive_learning} has feature blocks for both small object and larger objects that are output from the image encoder, and to better exploit the potential of ICCL, we optimise the input to the ICCL network. Specifically, the original feature blocks of the larger objects are modified to scale complementary semantic feature blocks, while the feature blocks of the small object are still retained as the output of the image encoder. This is because, our proposed ICCL is inter-scale contrastive complementary learning, which uses larger objects to guide the learning of the small object, and we adopt the more comprehensive feature blocks of larger objects as a teacher for guidance, and the learning of the small object will be better. The small object feature blocks still retain the feature blocks that output from the image encoder is because the optimisation of the contrastive complementary learning aims to make the classification branch better at learning small objects. In detail, the contrastive complement network branch only works in the training phase, and its purpose is to use the guidance of the ground category truths and the guidance of the large objects to form a more perfect representation of the small objects, which in turn can be used to optimise the representation of the small objects in the original classification branch, so that the classification branch can achieve a better representation of the small objects in the inference process, as shown in \autoref{fig:contrastive}. If we adopt a more perfect smaller objects' feature blocks that output from the CSLC as input of the contrastive complement network, the guidance of the large objects to the small objects in the contrastive complement network will be weaker, because the small objects' feature blocks are more perfect, and the space to be optimised by the contrastive complement network during the training process is smaller, and the guidance to the classification branch is also weaker, which ultimately leads to a weak optimal optimisation effect of the classification branch. Therefore, we retain the original inputs of the small objects, and in addition, we can maintain the consistency of the training and the inference.

%Implicit loop supervision. 

%\subsection{Scale-supervised loss function for proposals}

%Such an intra-class inter-scale optimisation approach is actually a looping optimisation approach throughout the model's multi-iters training, as shown in Figure. 

Through the collaborative approach described above, we better exploit the potential of complementary learning in UAV imagery to achieve better detection.

\subsubsection{End-to-end detection model}
\label{end-to-end}
We insert the components we propose above into the existing deep feature learning-based detection model to form an end-to-end detection model, named scale-robust complementary learning network (SCLNet). The whole model consists of two tasks, the object detection task, and the scale complementary task, unlike the general multi-task paradigm \cite{ding2020multi}, the scale complementary task is added to assist the object detection task, which further improves object detection performance. Inspired by the multi-task paradigm, we jointly optimise the two tasks together.
The loss function about object detection task $L_{detect}$ can be formulated as:
\begin{equation}
\begin{aligned}
L_{detect} &= L_{cls} + L_{contra} + L_{reg}\\
\end{aligned}
\end{equation}
where $L_{cls}$ and $L_{reg}$ indicate the loss function of the classification task and regression task in object detection. $L_{contra}$ denotes the loss function of contrastive labels and ground-truth, calculated in the same way as the $L_{cls}$, which detailed in \autoref{func:contra}.
The overall loss function $L_{total}$ for model training can be formulated as below:
\begin{equation}
\label{loss_total}
\begin{aligned}
L_{total} = \lambda_{comple}\cdot L_{comple}+ \lambda_{detect}\cdot L_{detect}\\
\end{aligned}
\end{equation}
where $\lambda_{comple}$ and $\lambda_{detect}$ respectively indicate the weights used to balance the two corresponding tasks. $L_{comple}$ indicates the scale-complementary loss function, which is the same as $L_{comple}$ described in \autoref{sec:complementary_loss}.
 
The way these three components work together can come to elaborate on the summary. The first two components, CSCL and ICCL, are used as core components to enhance the representation of objects of all scales in UAV images possessing scale variations of each category, especially for small objects, in order to obtain the desired robust representation capability, i.e., the semantic features extracted by the model are able to support the decoding of the downstream target detection task. The ECoop component, on the other hand, explores ways of suitable collaboration between components and between components and models in the object detection model framework in pursuit of better decoding for better detection performance. The above working mechanism culminates in an end-to-end UAV image object detection model.

\begin{table}
\footnotesize
\renewcommand{\arraystretch}{1.5}
\centering
\caption{Comparison of object scale variation in different scenario datasets.}
\label{tab:datasets}
\begin{tabular}{c|c|c|c}
\toprule
\multirow{2}{*}{scene} & \multirow{2}{*}{datasets} &\multirow{2}{*}{\makecell[c]{scale variation\\$<$2x}} &\multirow{2}{*}{\makecell[c]{scale variation\\$>$2x}} \\  
 & & & \\                            
\midrule
Natural scene & COCO\cite{lin2014microsoft}    &  53.2    & 46.8         \\ \cline{1-4}
Remote sensing  & FAIR1M\cite{sun2022fair1m}   & 79.4   &20.6           \\ \cline{1-4}
\multirow{2}{*}{UAV} & Visdrone\cite{du2019visdrone}       & 4.0    & 96.0      \\ \cline{2-4}
 & UAVDT\cite{du2018unmanned}    &17.6    &82.4         \\
\bottomrule
\end{tabular}
\end{table} 
\section{Experiments}
In this section, to evaluate the effectiveness of our approach, the paper constructs experiments from several perspectives. It is true that an approach designed for scale challenges is not only effective for a particular scale objects, but also improves the overall data detection performance. Therefore, our evaluation experiments focus on two main perspectives: the overall dataset and the objects of different scales included in it.

\subsection{Datasets}
For an ideal detection model, it should be robust to data with different degrees of scale challenges. We adopt two UAV image object detection datasets to build the evaluation experiments, VisDrone dataset\cite{du2019visdrone} and UAVDT dataset\cite{du2018unmanned}. As shown in \autoref{tab:datasets}, we count the scale variation of these two datasets and other scene datasets for comparative analysis.
According to the statistical results, the scale variation of UAV scene is much more significant than other scenes, such as natural scenes and general remote sensing scenes. The proportion of images with object scale variation greater than 2x in the two UAV datasets is 96.0\% and 82.6\% respectively, while the natural scene object detection dataset COCO\cite{lin2014microsoft} is 46.8\%, and the general remote sensing scene object dataset FAIR1M\cite{sun2022fair1m} is only 20.6\%. The comparison results show that the scale variation problem of the UAV scene is very significant.

VisDrone. Visdrone dataset\cite{du2019visdrone} is a large benchmark for object detection task and other computational vision tasks in UAV images. The dataset provides UAV aerial images and manual annotations. The image data is collected from 14 cities and villages in different regions, covering different weather and lighting conditions. Overall, the dataset is representative. The dataset is relatively large, including 10209 UAV images, of which the training set contains 6471, the validation set contains 548, and the testing set contains 3190. The manual annotations provided by the dataset cover ten predefined categories, namely pedestrian, bicycle, tricycle, person, truck, car, bus, van, motor, and awning-tricycle. Due to challenging factors such as occlusion, small objects, uneven distribution, and scale variation, the object detection task of this dataset is still difficult. In particular, scale challenges such as scale variations are prominent in this dataset. As shown in the statistical results in \autoref{tab:datasets}, the proportion of images in this dataset with scale variations greater than 2x is as high as 96.0\%. These scale problems seriously affect the detection of target instances contained in the dataset. Combining the above factors, most of the experiments are based on this dataset in order to pursue more credibility. As with most similar work, since the test set is not publicly available, this paper employs the training set to train the models in this paper, and the validation set to serve as the proof benchmark for both qualitative and quantitative experiments.The division of the training and validation sets follows the original provided images and annotations, which ensures that the experiments are fair. In summary, the 10,000 plus image size, the diversity of acquisition conditions, the labelling of ten fine-grained categories, and the 96\% of images with scale variations are enough characteristics of the dataset to support the evaluation of the effectiveness of the method proposed in this paper.

UAVDT. UAVDT dataset\cite{du2018unmanned} is another popular large benchmark for UAV image object detection. 
The dataset contains more than 40,000 UAV aerial images, covering different weather, different heights, and angles of UAVs. The manual annotations included are three predefined categories, namely car, bus, and truck. Scale challenges such as scale variation in this dataset are not as prominent compared to the Visdrone dataset. As shown in the statistical results in \autoref{tab:datasets}, although the proportion of images with scale variation more than 2x is much higher than that of other scene datasets (COCO dataset: 46.8\%, FAIR1M dataset: 20.6\%), reaching 82.4\%, it is slightly lower than the Visdrone dataset 96.0\%, it is a medium scale challenge dataset. In view of the above factors, we only compare with other state-of-the-art works on this dataset to illustrate the robustness of our proposed approach to datasets with different degrees of scale challenges. In summary, with a 40,000 plus image size, diversity of acquisition conditions, three fine-grained categories of annotation, and 86\% of images with scale variations, the characteristics of the dataset are also sufficient to support the comparative evaluation of the method proposed in this paper.

\subsection{Implement details}
Baseline model:
Most of the experiments are performed in the mmdetection framework\cite{chen2019mmdetection}.
Cascade RCNN\cite{cai2018cascade} is adopted as the baseline model of our proposed approach. The reason for this choice is that the cascade design of the detection head and the two-stage detection process of Cascade RCNN is a better baseline for the implementation of our proposed approach. The paper embeds the designed complementary learning components into the detection model, and the two-stage detection model is better interpretable\cite{zhou2018interpreting} and more conducive to explicit modelling scale challenges as we expect, thus resulting in an end-to-end model that is more interpretable.

Our model: Our SCLNet is an end-to-end detection model based on the complementary learning and the baseline model. The whole model network consists of one image encoder and two complementary learning implementations: comprehensive-scale complementary learning (CSCL) and inter-scale contrastive complementary learning (ICCL). Based on the whole complementary learning detection network, an end-to-end cooperation (Ecoop) is designed by several methods such as the cooperation of CSCL and ICCL, which further explore the potential of each component to achieve better detection performance. 

Training phase and testing phase: For the training of the model on the two UAV image object detection datasets, multi-scale training is used with input image size \{1024*765, 1360*800, 1024*1024\}. The size of the input image during the test phase is set to 1360*800. The baseline model and our SCLNet are both trained on 2 GPUs with 20 epochs and the SGD optimizer.  The sizes of the multiple large-kernel convolutions in \autoref{fig:model_all} are in order \{3, 5, 7, 11\} respectively. The reason for such a setup is that the combination of several convolution kernels of different sizes and several multi-scale feature maps with downsampling can cover the vast majority of target objects in UAV images. The scale difference between the largest scale feature map and the smallest scale feature map is a factor of 8. Thus, the range of the largest and smallest scales covered by the convolution kernel in such a pyramidal feature map is 88/3 times, and such a range can cover almost all objects. The number of proposals $N$ is preset to 500. For this set, we check the experimental settings of some advanced object detection works \cite{cai2018cascade, ma2023scale}, which are generally preset to 100 for natural scenes, but for UAV images, due to the larger number of objects, so in order to ensure that we can guarantee sufficient proposals, generally preset to 500, which is the setting in many advanced works\cite{deng2020global, yu2021towards}, and we also comply with such a preset. This also ensures a fair comparison. The $\lambda_{comple}$ and $\lambda_{detect}$ in \autoref{loss_total} are \{1.0, 0.3\} respectively, and the determined experiments are detailed in \autoref{sec:abla}. The initial learning rate is 0.0025, and the momentum parameter is 0.9. It is worth mentioning that the experiments of our method in the paper do not use any tricks, such as multi-scale testing and data augmentation, which are more conducive to evaluating the effectiveness of our proposed method, and at the same time, the experiments are more fair.

\subsection{Evaluation metrics}
The paper adopts two types of evaluation metrics for different purposes.

As with most object detection works\cite{cai2018cascade, ma2023scale}, we adopt $AP$, $AP_{50}$, and $AP_{75}$ as key evaluation metrics, which are complete and most popular for object detection evaluation experiments. $AP$, $AP_{50}$, and $AP_{75}$ indicate the mean averaged precision of all APs with IoU thresholds ranging from 0.5 to 0.95 with an interval of 0.05, APs at the 0.50 IoU threshold, and APs at the 0.75 IoU threshold respectively. 
Because the $AP$ metrics can reflect both prediction and recall situations, the $AP$ values calculated for the overall dataset can reflect the detection performance of the overall dataset. 
The larger the $AP$, $AP_{50}$, $AP_{75}$ value, the better the detection performance.
$AP$, or $mAP$, makes the average of the $APs$ of all categories, and the $AP$ of each category is calculated as the area under the Precision-recall curve. For the detailed calculation of $AP$, $N$ samples from each of the $C$ categories are assumed to have $Q$ positive examples, each of which corresponds to a recall value $(1/Q, 2/Q, \cdots, 1)$, and for each recall the maximum accuracy $P_{q}^{c}$ is calculated and then averaged over these $P_{q}^{c}$ values, which can also be formulated as:
\begin{equation}
\begin{aligned}
AP = \frac{1}{C}\sum_{c=1}^{C}\left (\frac{1}{Q}\sum_{q=1}^{Q} P_{q}^{c} \right ) \\
\end{aligned}
\end{equation}

In addition, we use evaluation metrics such as $AP_{l}$, $AP_{m}$, $AP_{s}$, $AR_{l}$, $AR_{m}$, and $AR_{s}$ to evaluate the detection performance of objects of different scales. $AP_{l}$, $AP_{m}$, and $AP_{s}$ can respectively indicate the comprehensive detection performance of large objects, medium objects, and small objects contained in the overall dataset. $AR_{l}$, $AR_{m}$, and $AR_{s}$ can respectively indicate the recall of large objects, medium objects, and small objects contained in the overall dataset. The Larger the    
$AP_{l}$, $AP_{m}$, $AP_{s}$, $AR_{l}$, $AR_{m}$, and $AR_{s}$ the better the detection performance in different scales. 

The detection of false alarms refers to the situation where the model mistakes the background for the foreground objects. For the quantitative evaluation metric false alarm rate $falsealarmrate$, which can be formulated as: 
\begin{equation}
\begin{aligned}
falsealarmrate = \frac{FP}{TP+FP} \\
\end{aligned}
\end{equation}
where $FP$ denotes the number of false alarm detection results. The $TP$ denotes the number of true detection results. The $FP+TP$ denotes the total detection results.

\begin{figure*}[t]
\centering
\includegraphics[width=6.7in]{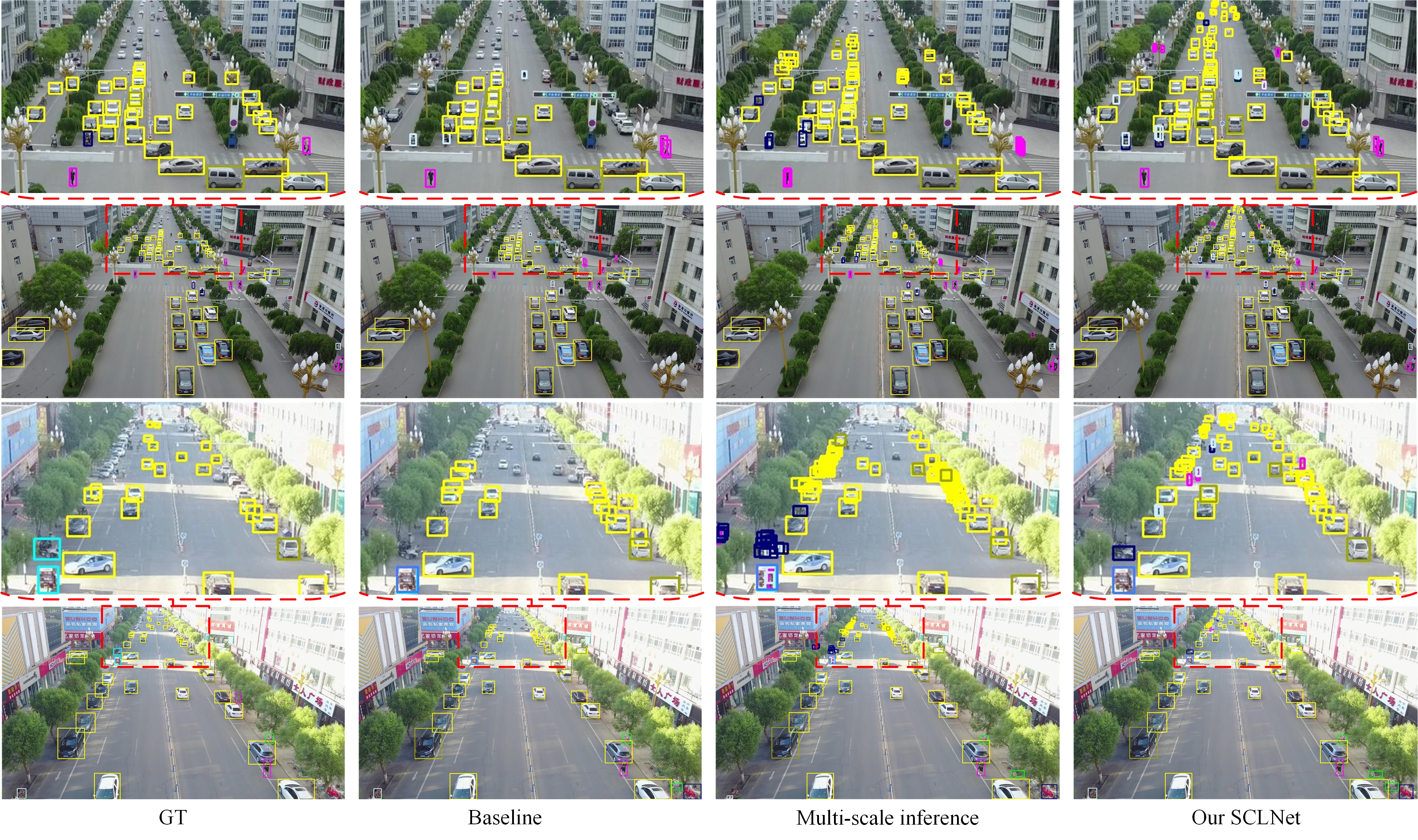}
\caption{The comparison of detection visualization results on Visdrone dataset among baseline model, multi-scale inference trick, and our SCLNet. The first and third rows are enlargements of local regions in the second and fourth rows, respectively. Our approach is able to effectively recall objects of different scales without bringing false alarms and possesses better robustness.}
\label{fig:com_vis_detect}
\end{figure*}
\subsection{Visualization of detection results}
As shown in \autoref{fig:com_vis_detect}, we visualize the ground-truth and detection results of baseline models, multi-scale inference, and our proposed approach for qualitative comparative analysis to illustrate the effectiveness of our proposed approach. It is easy to observe that the baseline model (second column) performs better for object detection at larger scales, but for objects at smaller scales, the model lacks sufficient perception ability and recall is poor. For the popular multi-scale inference trick (third column), while such a trick can recall more objects of different scales, it brings a large number of false alarms. Our SCLNet can recall more objects of different scales with almost no false alarms from the qualitative observation perspective, especially for small objects, which indicates that our model is better at detecting objects of all scales. To better demonstrate the effectiveness of our proposed SCLNet for reducing false alarm detection, we establish a quantitative comparison among baseline model, multi-scale inference and our SCLNet in \autoref{tab:falsealarm}. According to the experimental results in the table, our SCLNet reduces the false alarm rate from 10.6\% to 8.5\% compared to baseline, and reduces 6.7\% compared to multiscale inference. Both qualitative observations and quantitative comparisons are sufficient to prove that our proposed SCLNet can indeed reduce false alarms. In addition, the detection results of our SCLNet contain some objects not labelled in ground-truth, which are actually present in UAV images, which demonstrates the robustness of our approach, although this is harmful to the computation of the evaluation metrics. In summary, it can be qualitatively demonstrated that our approach is able to effectively address the scale challenges in UAV image object detection and is a more robust method with better detection performance.
\begin{table}
\footnotesize
\renewcommand{\arraystretch}{1.5}
\centering
\caption{Comparison of false alarm rates on the Visdrone validation set among baseline model, multi-scale inference and our proposed SCLNet.}
\label{tab:falsealarm}
\begin{tabular}{c|c|c|c}
\toprule                          
Method  &Baseline    & Multi-scale inference & Our SCLNet \\\cline{1-4}
Flase alarm rate  &10.6 &15.2 &8.5                     \\
\bottomrule
\end{tabular}
\end{table}

%High recall to small object. 

%Robust to all scales. 

\subsection{Comparison to other state-of-the-art models}
In this section, we construct quantitative comparisons of our proposed approach with some of the state-of-the-art works of recent years in two UAV image object detection datasets, as shown in \autoref{tab:visdrone} and \autoref{tab:UAVDT}. Through the quantitative comparative analyses, we aim to demonstrate that our approach is a competitive model in UAV image object detection.
\begin{table}
\footnotesize
\renewcommand{\arraystretch}{1.5}
\centering
\caption{Comparison among other state-of-the-art models and our models on the validation set of the Visdrone dataset. '-' indicates that the corresponding value is not reported in the relevant paper.}
\label{tab:visdrone}
\begin{tabular}{c|c|c|c|c}
\toprule
 Method & Backbone & $AP$ & $AP_{50}$ & $AP_{75}$  \\                              
\midrule
CascadeRCNN\cite{cai2018cascade}(CVPR2018)         &ResNet50  & 27.5     & 46.0     & 28.6     \\ \cline{1-1}
ClusDet\cite{yang2019clustered}(ICCV2019)      &ResNeXt101  & 29.8     & 51.9     & 29.9     \\ \cline{1-1}
DMNet\cite{li2020density}(CVPR2020)         &ResNeXt101   &29.4       &  49.3   & 30.6      \\ \cline{1-1}
GLSAN\cite{deng2020global}(TIP2020)  &ResNet50  &30.7       &\textbf{55.4}    &30.0        \\ \cline{1-1}
Swin\cite{liu2021swin}(ICCV2021)  & Swin    & 29.1 &- & - \\ \cline{1-1}
DSHNet\cite{yu2021towards}(WACV2021) &ResNet50  &30.3    & 51.8   & 30.9  \\   \cline{1-1}
LSEM\cite{kong2022realizing}(NC2022)  &ResNet50  & 28.0     & 44.9   & -      \\ \cline{1-1}
QueryDet\cite{yang2022querydet}(CVPR2022) &ResNet50  & 28.3      & 48.1   & 28.7       \\ \cline{1-1}
Lit\cite{pan2022less}(AAAI2023)           &  LIT     &  27.2     & 45.7   & 28.1  \\ \cline{1-1}
ToMe\cite{bolya2023token}(CVPR2023)       &  TOME     &  27.8     & 46.9   & 28.5  \\ \cline{1-1}
DeformPM\cite{chen2021dpt}(ACM MM 2021)   &  DPT     &  29.2     & 49.1   & 30.3  \\ \cline{1-1}
CEASC\cite{du2023adaptive}(CVPR2023)       & ResNet50 &  20.8    & 35.0     & 27.7        \\  \cline{1-1}
SDPNet\cite{ma2023scale}(TGRS2023)       & ResNet50 &  30.2    & 52.5     & 30.6        \\   \midrule
Ours     & ResNet50  &\textbf{31.1}       &52.9    & \textbf{31.5}        \\
\bottomrule
\end{tabular}
\end{table}

Visdrone. The quantitative results of the detection performance of some state-of-the-art works and our proposed approach on the Visdrone dataset are listed in \autoref{tab:visdrone}. Compared with the general detection model Cascade-RCNN\cite{cai2018cascade}, our proposed approach outperforms $AP$:3.6\%, $AP_{50}$:6.9\%, $AP_{75}$:2.9\% in the case of the same backbone network resnet50. Compared with the UAV image object detection models ClusDet\cite{yang2019clustered} and DMNet\cite{li2020density}, our proposed approach is higher $AP$:1.3\%, $AP_{50}$:1.0\%, $AP_{75}$:1.6\% and $AP$:1.7\%, $AP_{50}$:3.6\%, $AP_{75}$:0.9\%, respectively, and the backbone network used by our proposed method is resnet50, while the backbone network used by these two models is resnext101. Since the number of parameters and computation of the backbone network is a high percentage of the model, our method achieves a high detection accuracy with a significant reduction in the number of parameters and computation. For GLSAN\cite{deng2020global}, which also addresses scale challenges, our proposed method outperforms $AP$:0.4\%, $AP_{75}$:1.5\%. For the latest work CEASC\cite{du2023adaptive}, our proposed approach performs superiorly, exceeding $AP$: 10.3\%, $AP_{50}$:17.9\%, $AP_{75}$:3.8\%. For DSHNet\cite{yu2021towards} and LSEM\cite{kong2022realizing}, which are two works to optimise the detection head, our proposed approach outperforms these two models respectively by $AP$:0.8\%, $AP_{50}$:1.1\%, $AP_{75}$:0.6\% and $AP$:3.1\%, $AP_{50}$:8.0\%. For transformer-based works Swin\cite{liu2021swin} and QueryDet\cite{yang2022querydet}, our proposed approach with the same backbone network resnet50 is higher than these two works by $AP$:2.0\% and $AP$:2.8\%, $AP_{50}$:4.8\%, $AP_{75}$:2.8\% respectively. For some other advanced transformer-based works, Lit\cite{pan2022less}, ToMe\cite{bolya2023token}, and DeformPM\cite{chen2021dpt}, improvements have been proposed from different perspectives, and our proposed method outperforms their detection performance.
\begin{table}
\footnotesize
\renewcommand{\arraystretch}{1.5}
\centering
\caption{Comparison among other state-of-the-art models and our models on the test set of the UAVDT dataset. '-' indicates that the corresponding value is not reported in the relevant paper.}
\label{tab:UAVDT}
\begin{tabular}{c|c|c|c|c}
\toprule
 Method & Backbone & $AP$ & $AP_{50}$ & $AP_{75}$  \\                              
\midrule
%Cascade RCNN(CVPR2018)         &ResNet50  &    &      &    \\ \cline{1-1}
ClusDet\cite{yang2019clustered}(ICCV2019)        &ResNet50   & 13.7      &26.5     &12.5      \\ \cline{1-1}
DREN\cite{zhang2019fully}(ICCV2019)  &ResNet101  & 17.1    & -   & -  \\ \cline{1-1}
GLSAN\cite{deng2020global}(TIP2020)  &ResNet50  & 17.0      &28.1    &18.8        \\ \cline{1-1}
AMRNet\cite{wei2020amrnet}(2020)  &ResNet50       &18.2   &30.4   & 19.8 \\ \cline{1-1}
DSHNet\cite{yu2021towards}(WACV2021)        &ResNet50   &17.8       &30.4     &19.7      \\ \cline{1-1}
CDMNet\cite{duan2021coarse}(ICCV2021)  &ResNet50   &16.8    &29.1    &18.5       \\ \cline{1-1}
%OGMN\cite{li2023ogmn}(ISPRS2023)       & ResNet50 &20.9      &34.5     &23.2       \\ \cline{1-1}
CEASC\cite{du2023adaptive}(CVPR2023) &ResNet50  &17.1   &30.9  & 17.8 \\ \midrule
Ours     & ResNet50  & \textbf{20.0}      &\textbf{33.1}    & \textbf{22.3}       \\
\bottomrule
\end{tabular}
\end{table}

UAVDT. As shown in \autoref{tab:UAVDT}, the quantitative results of the detection performance of some state-of-the-art works and our proposed approach on the UAVDT dataset are listed. For DREN\cite{zhang2019fully} using the resnet101 backbone network, although our proposed method uses a lightweight backbone network resnet50, our method is superior to this work by $AP$:2.9\%. Compared to DSHNet\cite{yu2021towards} of optimising the detection head, our proposed approach goes beyond $AP$:2.2\%, $AP_{75}$:2.6\%. For UAV image detectors ClusDet, GLSAN, AMRNet, and CDMNet, our proposed approach performs better in detection performance. The approach we proposed has higher $AP$:6.3\%, $AP_{50}$:6.6\%, $AP_{75}$:9.8\% than ClusDet\cite{yang2019clustered}; higher $AP$:3.0\%, $AP_{50}$:5.0\%, $AP_{75}$:3.5\% than GLSAN\cite{deng2020global}; higher $AP$:1.8\%, $AP_{75}$:2.5\% than AMRNet\cite{wei2020amrnet}; higher $AP$:3.2\%, $AP_{50}$:4.0\%, $AP_{75}$:3.8\% than CDMNet\cite{duan2021coarse}. Compared to latest work CEASC\cite{du2023adaptive}, our proposed approach is higher than this work by $AP$:2.9\%, $AP_{75}$:4.5\%.  

\begin{figure}[t]
\centering
\includegraphics[width=3.4in]{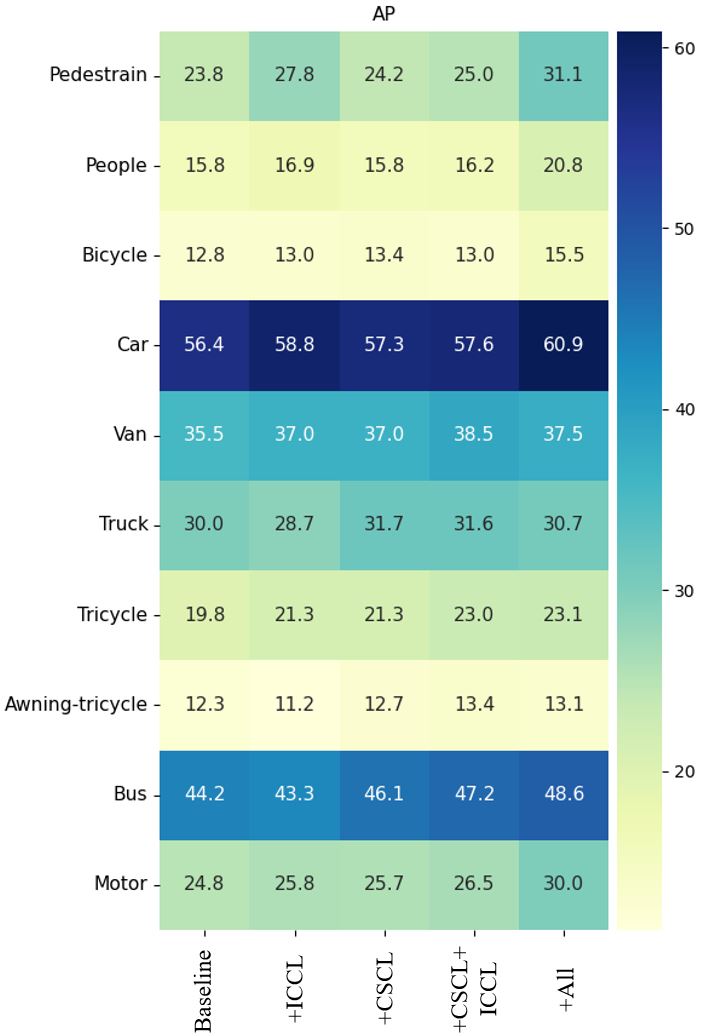}
\caption{The qualitative results of ablation experiments for all fine-grained categories on the validation set of Visdrone dataset. Darker colours indicate higher AP values and better detection performance. '+All' means that all components are added.}
\label{fig:class_ap}
\end{figure}

Room for improvement. For some advanced works such as TPH-YOLOv5++\cite{zhao2023tph} and Cascade Zoom\cite{meethal2023cascaded}, they adopt some tricks not used in the experiments of our proposed approach in this paper. TPH-YOLOv5++ employs the MixUp, Mosaic, Flip and Rotate. Cascaded Zoom training augmentation approach with density crop and multi-stage inference approach are adopted. Given the fairness of the comparison, we did not list these jobs for comparison, but we will focus on the merits of these works in the future study of algorithmic landing.  The UFPMP-Det\cite{huang2022ufpmp} is indeed an excellent one, cleverly using the bag-of-instance-words approach to address the object distribution, and also achieving good detection performance, where our approach does end up achieving slightly less accuracy than this work. However, in this paper we propose an approach that aims to address the problem of significant scale variations by proposing two complementary learning ideas to address this problem, and experiments such as ablation experiments demonstrate the effectiveness of our proposed approach. And in terms of inference speed, the inference speed of our proposed method is 0.108s, the inference speed of this method is 0.152s, our proposed method is 40.7\% faster, faster inference speed is also important for practical applications. In future practical applications, integrating our proposed method with these advanced methods such as UFPMP-Det into a parallel modelling framework might achieve even more stunning detection performance. The experiments in this paper focus on proving the effectiveness of our proposed method, and these specific algorithmic landings will continue to be investigated and the code will be open-sourced in the future. At the same time, the deployment details will also be open sourced.

Based on the above analysis, no matter for the general detection model based on CNN, the detection model based on transformer, or the recent UAV detection methods, the quantitative detection performance of our proposed approach on the two UAV object detection datasets is superior to these methods. Although there is still room for improvement in terms of practical applications, it can be demonstrated that our approach is competitive in UAV image object detection task.

\subsection{Ablation experiments of the different components}
\label{sec:abla}
\begin{figure*}[t]
\centering
\subfloat[stage 0]{
\includegraphics[scale=0.375]{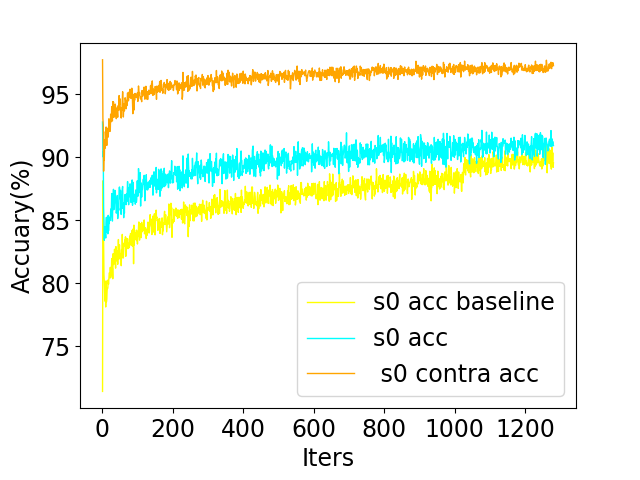}} 
\subfloat[stage 1]{
\includegraphics[scale=0.375]{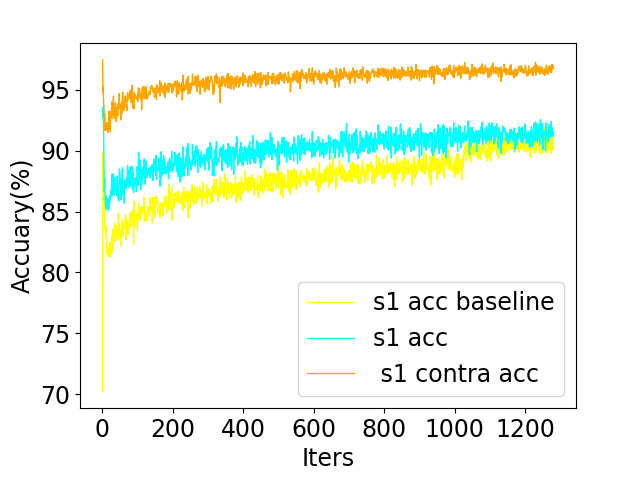}} 
\subfloat[stage 2]{
\includegraphics[scale=0.375]{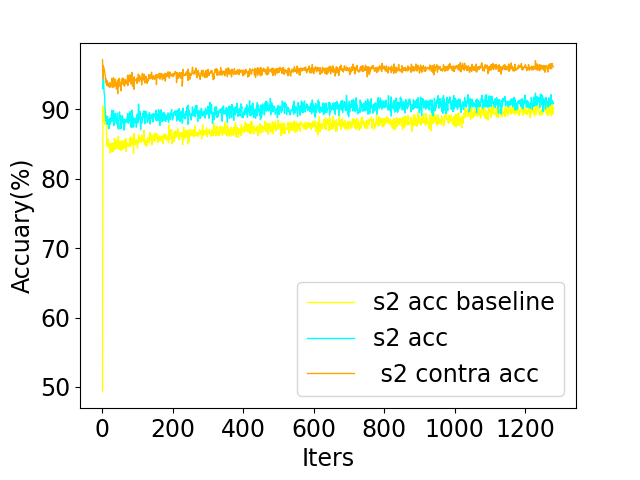}}
\caption{Comparison of accuracy convergence curve in different cascade stages. The three curves in each subfigure represent the accuracy of the baseline model, the accuracy of the classification branch of our model, and the accuracy of the contrastive complement network branch. Our proposed contrastive complementary network branch achieves much higher accuracy than the baseline model and also improves the accuracy of the classification branch in our model.}
\label{fig_accuary}
\end{figure*}

To demonstrate the effectiveness of each component in our approach, this section set up the ablation experiments on the Visdrone dataset in this subsection. 
For the purpose of comprehensive evaluation, we conduct quantitative analysis from multiple evaluation perspectives of the whole dataset. 
\autoref{tab:ablation} lists the numerical results of the ablation experiments on the overall data. And the \autoref{fig:class_ap} illustrates ablation experiments results of multiple fine-grained categories. The comparison of the convergence curves of the accuracy during the training process is presented in \autoref{fig_accuary}. 
\begin{table*}
\footnotesize
\renewcommand{\arraystretch}{1.5}
\centering
\caption{Ablation accuracy comparison among the baseline model, and models with the different components in our approach on the validation set of the Visdrone dataset. The ECoop is dependent on CSCL and ICCL components.}
\label{tab:ablation}
\begin{tabular}{c|c|c|c|c|c|c|c|c|c|c|c}
\toprule
\multirow{2}{*}{CSCL}  &\multirow{2}{*}{ICCL}  &\multirow{2}{*}{ECoop} &\multirow{2}{*}{$AP$} &\multirow{2}{*}{$AP_{50}$} &\multirow{2}{*}{$AP_{75}$}&\multirow{2}{*}{$AP_{s}$}&\multirow{2}{*}{$AP_{m}$}&\multirow{2}{*}{$AP_{l}$}&\multirow{2}{*}{$AR_{s}$}&\multirow{2}{*}{$AR_{m}$}&\multirow{2}{*}{$AR_{l}$} \\
                                        & & &   & & & &  & &  &  &  \\ 
\midrule
          &         &         &27.5         &46.0          &28.6          &17.3          &40.0          &46.7          &31.1           &57.4          &61.7     \\
  &  $\surd$        &         &28.3         &50.4          &28.2          &21.2         &38.5         &37.9          &37.8  &55.2          &51.9   \\ 
$\surd$           & &   	  &28.5         &46.7          &29.7          &18.2          &41.3          &47.4          &32.8           &57.7          &60.3     \\
 $\surd$  & $\surd$ &   	  &29.2         &47.0          &30.9          &18.6          &\textbf{43.0}          &\textbf{49.0}          &33.2           &\textbf{59.7}          &\textbf{63.1}   \\
$\surd$  & $\surd$ & $\surd$ &\textbf{31.1}   &\textbf{52.9}   &\textbf{31.5}   &\textbf{22.4}   &41.4   &47.4    &\textbf{39.4}     &57.7  &62.0 \\
\bottomrule
\end{tabular}
\end{table*}

\textbf{Effect of the comprehensive-scale complementary learning (CSCL).}
One of our proposed implementations of complementary learning for UAV image object detection is comprehensive-scale complementary learning (CSCL). The third row in \autoref{tab:ablation} shows the experimental results with only the addition of CSCL component, compared to the baseline model without any additional components in the first row, our proposed model achieves a boost of $AP$:1.0\%, $AP_{50}$:0.7\%,$AP_{75}$:1.1\%, which is brought about by the CSCL on all the three scales of the objects set, namely, the large objects set $AP_{l}$:0.7\%, the medium objects set $AP_{m}$:1.3\%, $AR_{m}$:0.3\%, the small objects set: $AP_{s}$:0.9\%, $AR_{s}$:1.7\%. The CSCL component is originally designed to enhance the representation of all scales in a complementary way. The detection accuracy of the object set at small, medium and large scales is improved by the addition of the CSCL component, which proves that the original design intention of the CSCL is realised and that the scale robustness of the detection is indeed enhanced.
The third column in the fine-grained category ablation results in \autoref{fig:class_ap} adds the CSCL component compared to the first column, and the $AP$ accuracy is improved for all ten categories.
In \autoref{tab:CSCL}, we establish additional experiments by adding CSCL to the Faster R-CNN, YOLO, and DETR models to evaluate the effectiveness of the CSCL module. What can be observed from \autoref{tab:CSCL} is that the detection accuracy is improved with the addition of the CSCL component based on the three models, with APs improved by 3.4\%, 2.2\%, and 3.1\%, respectively, which proves that our proposed CSCL component is effective and can indeed improve the detection performance.
Thus, the overall data improvement, the improvement on three different scales, and the improvement on all fine-grained categories demonstrate the effectiveness of our proposed CSCL component, and the implementation of this component indeed achieves a better detection performance on the comprehensive scales. 

\begin{table}
\footnotesize
\renewcommand{\arraystretch}{1.5}
\centering
\caption{The ablation study about comprehensive-scale complementary learning (CSCL) component.}
\label{tab:CSCL}
\begin{tabular}{c|c|c|c}
\toprule
 Method & $AP$ & $AP_{50}$ & $AP_{75}$  \\                              
\midrule
Faster R-CNN\cite{ren2015faster}       &20.7  &36.4   &20.5  \\ \cline{1-1}
Faster R-CNN+CSCL  &24.1  &45.2   &22.7   \\ \cline{1-4}
YOLO\cite{redmon2016you}       &21.7  &45.4   &18.3  \\ \cline{1-1}
YOLO+CSCL  &23.9  &48.1   &21.3   \\ \cline{1-4}
Deformable DETR\cite{zhu2020deformable}       &18.2  &35.2   &17.0  \\ \cline{1-1}
Deformable DETR+CSCL  &21.3  &39.7   &20.3   \\ 
\bottomrule
\end{tabular}
\end{table}

\textbf{Effect of the inter-scale contrastive complementary learning (ICCL).}
Another implementation of complementary learning for UAV image object detection that we propose is inter-scale contrastive complement learning (ICCL). The results in the second row of \autoref{tab:ablation} with the addition of the ICCL component achieve a boost of $AP$:0.8\%, $AP_{50}$:4.4\%, $AP_{75}$:1.1\% in the overall data compared to the first row, and $AP_{s}$:3.9\%, $AR_{s}$:1.7\% for the small objects set as well. The second column in \autoref{fig:class_ap} with the addition of the ICCL component represents a clear improvement in $AP$ accuracy compared to the first column for several categories that are predominantly small objects, especially for the pedestrian and car categories where the AP improvement reaches 4.0\% and 2.4\%, respectively. The addition of the ICCL component in \autoref{fig_accuary} results in a much higher accuracy (orange curve) than the baseline model (yellow curve), as well as achieving a much faster convergence rate, and also the contrastive complement learning method helps the classification branch (blue curve) to achieve the higher accuracy and is significant for all three stages of improvement in the cascade detection head. All in all, the improvement on the overall dataset, the improvement on the set of small objects, the significant improvement on the fine-grained categories dominated by small objectss, and the higher accuracy curves demonstrate that the complementary learning of large objects to small objects comparisons achieved by our proposed ICCL component is effective, and in particular, the improvement in the detection performance of the small objects is significant. All in all, the improvement on the overall dataset, the improvement in the small objects set, the significant improvement in the fine-grained categories dominated by small objects, and the higher accuracy curves demonstrate that the contrastive complementary learning of large objects to small objects achieved by our proposed ICCL component is effective, and in particular for small objects, the improvement in detection performance is significant.
\begin{figure*}[t]
\centering
\includegraphics[width=6.9in]{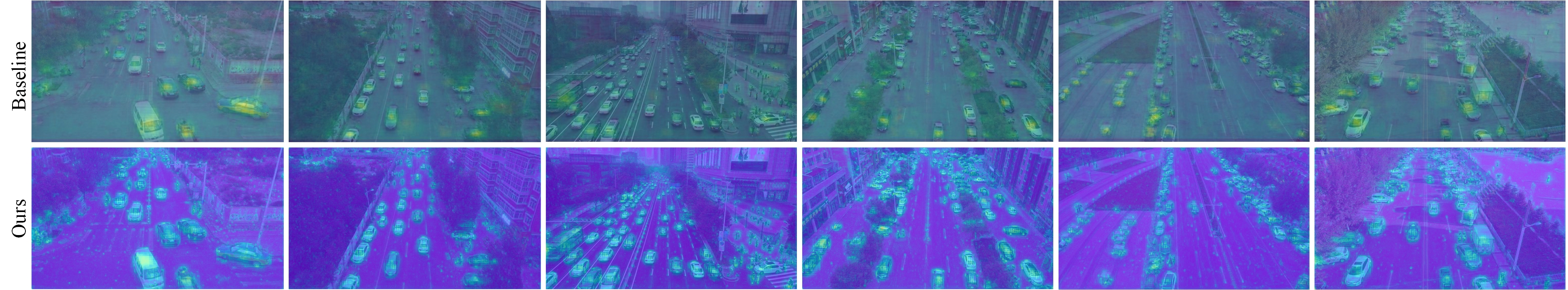}
\caption{The comparison of the feature maps among the baseline model and our approach. Our approach allows the model to perceive more objects at all scales, especially for small objects. At the same time, our approach is able to better distinguish between foreground and background compared to the baseline model.}
\label{fig:hotmap}
\end{figure*}

\begin{table}
\footnotesize
\renewcommand{\arraystretch}{1.5}
\centering
\caption{The ablation study about hyperparameters $\lambda_{comple}$ and $\lambda_{detect}$.}
\label{tab:hyperparameter}
\begin{tabular}{c|c|c|c|c|c|c}
\toprule
$\lambda_{comple}$      &0.7 &0.8  &0.9   &1.0 &1.1 &1.2  \\ \cline{1-1}
$AP$                    &30.8&31.0 &30.9  &\textbf{31.1}&31.0&31.0  \\ \cline{1-7}
$\lambda_{detect}$      &0.1  &0.2   &0.3  &0.4 &0.5 &0.6  \\ \cline{1-1}
$AP$                    &30.6 &30.9  &\textbf{31.1} &30.9&31.0&30.9 \\ 
\bottomrule
\end{tabular}
\end{table}

\begin{table}
\footnotesize
\renewcommand{\arraystretch}{1.5}
\centering
\caption{Comparison of with other advanced works for scale challenges on the validation set of the Visdrone dataset. '-' indicates that the corresponding value is not reported in the relevant paper.}
\label{tab:scale_challenges}
\begin{tabular}{c|c|c|c|c|c}
\toprule
 Method & Backbone & $AP$ & $AP_{s}$ & $AP_{m}$ & $AP_{l}$ \\                              
\midrule
CEASC\cite{du2023adaptive}  &ResNet50 & 20.8      &-     &-   &-       \\ \cline{1-1}
ClustDet\cite{yang2019clustered}  & ResneXt101  &28.4  &19.1 & 40.8 & 54.4 \\ \cline{1-1}
DMNet\cite{li2020density}  & ResNet50  & 29.4  & 21.6  & 41.0 & \textbf{56.9} \\ \cline{1-1}  
QueryDet\cite{yang2022querydet} &ResNet50   & 28.3      & 17.9     & 30.4   & 36.7   \\ \cline{1-1}
GLSAN\cite{deng2020global}  &ResNet50  & 30.7  & -   & -   & -    \\ \cline{1-1}
Ours     & ResNet50  & \textbf{31.1}   & \textbf{22.4}   & \textbf{41.4}  &  47.4    \\
\bottomrule
\end{tabular}
\end{table}

\textbf{Effect of the end-to-end cooperation (ECoop).}
The implementation of the end-to-end cooperation (ECOOP) component is dependent on CSCL and ICCL, as ECOOP is an exploitation of the potential of both components. The fifth row in \autoref{tab:ablation} achieves an improvement of $AP$:1.9\%, $AP_{50}$:5.9\%, $AP_{75}$:0.6\% in the overall data compared to the fourth row with the addition of the ECoop component. Compared to the baseline model in the first row, the lift on the overall dataset achieves $AP$:3.6\%, $AP_{50}$:6.9\%, $AP_{75}$:2.9\%, and $AP_{l}$:0.7\%, $AR_{l}$:0.3\% for the large objects set, $AP_{m}$:1.4\%, $AR_{m}$:0.3\% for the medium objects set, $AP_{s}$:5.1\%, $AR_{s}$:8.3\% for the small objects set. For the fine-grained results in \autoref{fig:class_ap}, the addition of Ecoop results in varying degrees of $AP$ improvement in the fine-grained categories, with the fifth column achieving $AP$ improvements of 6.1\%, 4.6\%, and 3.5\% for pedestrian, people, and motor categories, respectively, compared to the fourth column. Compared to the baseline model in the first column, our approach achieves improvements in all fine-grained categories, especially for the categories of pedestrian, motor, people, and car by 7.3\%, 5.2\%, 5.0\%, and 4.5\%. After the addition of ECoop, the accuracy at large and medium scales in the fifth row in \autoref{tab:ablation} decreases than the fourth is due to the fact that the ICCL component aims to improve the representation of small objects, whereas ECoop makes use of the features of larger objects extracted by CSCL to compensate for the lack of small object representation, and therefore has a negative impact on the representation of larger objects. However, in terms of overall performance, the AP in the fifth row is still significantly improved than the fourth row and the detection performance is improved for all scales, large, medium, and small, compared to the baseline model, all of which proves the effectiveness of our proposed method. It can be concluded that the Ecoop component is able to effectively exploit the potential of CSCL and ICCL, leading to a significant improvement in detection performance, both in terms of overall data and fine-grained categories.

\textbf{Effect of the hyperparameters $\lambda_{comple}$ and $\lambda_{detect}$.}
Hyperparameters $\lambda_{comple}$ and $\lambda_{detect}$ are provided for balancing different tasks in the training phase of multi-task co-optimisation. Their appropriately matched values determine whether the collaboration of different tasks is optimal or not, so the determination of the values is important. Therefore, we set up relevant hyperparameter ablation experiments, firstly, we preset the value of $\lambda_{detect}$ to be 0.5 provided that the model is able to converge, then, we adjust the value of $\lambda_{comple}$ to pursue the optimum, and $\lambda_{comple}$ is determined to be 1.0, and then, we preset the value of $\lambda_{comple}$ to be 1.0 to seek for the optimum value of $\lambda_{detect}$, and finally, $\lambda_{detect}$ is determined to be 0.3. Finally, the two hyperparameters $\lambda_{detect}$ and $\lambda_{comple}$ are determined to be 0.3 and 1.0.

In general, each component of our approach is effective and achieves the effects we expect, and the detection model they form cooperatively does achieve better detection performance.

\subsection{Analysis for scale challenges}
The original intention of our proposed method is to address the scale challenges in UAV image object detection. In order to better prove the effectiveness of our approach for scale challenges, in this subsection the qualitative and quantitative experiments demonstrate our proposed approach from several different perspectives.

\subsubsection{Quantitative comparison with other advanced methods for addressing scale challenges} 
As shown in \autoref{tab:scale_challenges}, we list the quantitative results of some recent advanced works for addressing scale challenges on the Visdrone dataset, and compare our proposed approach with these methods. For both DMNet and GLSAN addressing the objects scale variable distribution problem in UAV images, our approach outperforms $AP$:1.7\%, $AP$:0.4\% respectively compared to these two works using the same backbone.  For ClustDet with backbone ResneXt101, another work of addressing objects scale variable distribution problem, our approach outperforms $AP$:2.7\% in the case of using Resnet50 with much fewer parameters. For CEASC and QueryDet, which address the small objects problem, our approach outperforms $AP$:9.3\% and $AP$:2.8\%, respectively. In addition, for small objects and medium objects, our approach presents excellent performance. The above comparisons and analyses illustrate that our proposed method is a competitive approach for addressing scale challenges compared to existing works for addressing scale challenges in UAV image object detection.
\begin{table}
\footnotesize
\renewcommand{\arraystretch}{1.5}
\centering
\caption{Comparison of model complexity with other works on validation set of the Visdrone dataset.}
\label{tab:time}
\begin{tabular}{c|c|c|c|c|c}
\toprule                          
\multirow{2}*{method} &\multirow{2}*{\shortstack{Cascade\\RCNN\cite{cai2018cascade}}} & \multirow{2}*{\shortstack{RetinaNet\\\cite{lin2017focal}}} & \multirow{2}*{\shortstack{QueryDet\\\cite{yang2022querydet}}} & \multirow{2}*{\shortstack{GLSAN\\\cite{deng2020global}}} & \multirow{2}*{Ours} \\ 
  &  &  &  &   & \\ \midrule
%$AP$  &27.5 &26.2 &28.3 &30.7 &31.1                        \\
%$FPS$ & 12.3 &2.6 &2.8  &1.3  & 9.2                   \\
$AP$  &27.5 &18.2 &19.5 &30.7 &31.1                        \\
$FPS$ &12.3 &13.2 &14.0 &1.3  & 9.2                   \\
\bottomrule
\end{tabular}
\end{table}
\subsubsection{Qualitative analysis of feature maps}
The feature maps of the baseline model and our proposed approach are presented as hotmaps in \autoref{fig:hotmap}.
Compared to the baseline(first row), our approach (second row) is able to perceive more objects and activate more regions of object parts with the complementary way by CSCL component.
These evidences can prove that our proposed implementation of comprehensive-scale complementary learning is helpful to form a comprehensive and robust feature representation at all scales. 
Especially for small objects, we are not only able to perceive them and form more comprehensive features, which makes the detection of small objects more robust, thanks to two aspects, on the one hand, it is because the implementation of the comprehensive-scale complementary learning that we propose can increase the perceptual ability of small objects, and furthermore, it is because of the intra-category contrastive complementary learning that we propose, which enables the model to achieve a more robust perception of small objects that lack texture detail information. In addition, our proposed approach can better distinguish between foreground and background compared to the baseline model, the darker blue colour in \autoref{fig:hotmap} indicates that the model is considered more as background and the darker yellow colour indicates that the model is considered more as foreground, a lot of background pixels are misclassified as foreground in the baseline model with a certain probability, whereas our approach is able to keep the interference of the background noise very well, which makes the foreground's representational features more robust. Therefore, it can be demonstrated that our approach is more robust for UAV object detection containing scale challenges.

%\subsubsection{Qualitative visualization comparison of some instances at different scales}
%In order to better illustrate the robustness of our proposed approach for objects at different scales in UAV images, we list the visualisation results of the detection performance for some instances at different scales, as shown in \autoref{}.

To summarise, the qualitative or quantitative comparisons and analyses of the above different perspectives illustrate that our approach is effective for addressing the effectiveness of our approach for addressing the scale challenges.

\subsection{Analysis of complexity}
To evaluate that our approach can meet the real-time requirements of practical application scenarios, we construct a time-complexity evaluation experiment and compare it with other popular works in this section. 
The evaluation metric of inference speed $FPS$ can accurately project the time complexity of the model. Therefore, we adopt inference speed as a measure of comparison time complexity in \autoref{tab:time}. Compared to the generic two-stage object detection method Cascade RCNN, our proposed method achieves $FPS$:9.2, although slightly lower than its $FPS$:12.3 on speed, is significantly higher on accuracy with $AP$:31.1\% than its $AP$:27.5\%. Compared to the generic one-stage RetinaNet, although our proposed method is slightly slower in the inference speed, the detection accuracy is higher by $AP$12.9\%.
For the popular UAV image object  detection model GLSAN, our approach outperforms the inference speed by $FPS$:7.9, and the detection accuracy by $AP$:0.4\%. For the latest UAV image object detection model QueryDet, although our approach is slower in the inference speed, the detection accuracy is significantly higher by $AP$:11.6\%. The summary that can be drawn is that our method guarantees excellent detection accuracy with guaranteed detection speed, which is reliable for practical applications because both detection accuracy and inference speed are better. In general, our approach can meet the real-time requirements.

\subsection{Effect of the resolution}
In order to illustrate the effect of the resolution of the input image on the detection accuracy of the model based on the method proposed in this paper in a UAV image object detection scenario, we have additionally set up some experiments, the results of which are presented in \autoref{tab:resolution}. 
From the experimental results, the effect of resolution on detection accuracy is basically in line with the expected experience. However, when adjusting the resolution from 1360*800 to 1000*800, the accuracy of detection only decreased by 0.4\% and 0.2\%, which indicates that such a resolution adjustment does not have much effect on the detection performance of our proposed method. When increasing the resolution to 1920*1080, a resolution that is also used in some works, the detection accuracy is improved more significantly, by 1.4\% and 1.1\%, respectively. The above experiments demonstrate that our proposed method is still an effective object detection model for UAV images. In practical applications, the resolution of the actual input image is often determined based on the comprehensive consideration of the detection accuracy and inference speed.
\begin{table}
\footnotesize
\renewcommand{\arraystretch}{1.5}
\centering
\caption{Effect of input resolution on detection accuracy of the model based on our proposed approach.}
\label{tab:resolution}
\begin{tabular}{c|c|c|c}
\toprule                          
\multirow{2}*{Dataset}  & \multicolumn{3}{c}{Resolution}  \\\cline{2-4}
                        & 1000*800 & 1360*800 & 1920*1080 \\\cline{1-4}
Visdrone  &30.7 &31.1 &32.5                     \\\cline{1-4}
UAVDT     &19.8 &20.0 &21.1                   \\
\bottomrule
\end{tabular}
\end{table}

\section{Conlusion}
In the paper, we propose a scale-robust comprehensive learning network to improve UAV image detection performance. Two complementary learning implementations to address characteristics of scale challenges. We propose a scale complementary decoder and a scale complementary loss function to implement comprehensive-scale complementary learning for forming a comprehensive representation of all scale objects. We propose a contrastive complement network and a contrastive complement loss function to implement inter-scale contrastive complementary learning, making large objects with more texture detailed information to guide the learning of small objects. Additionally, we propose end-to-end cooperation between the two implementations and with the detection model to exploit the potential of each component. The evaluation experiments on Visdrone and UAVDT datasets prove the effectiveness, and robustness of our approach. In the future, we will introduce more complementary learning techniques from inter-modal, inter-temporal and other perspectives, such as depth information, to further improve the detection performance of UAV image object detection. The precise and high-resolution depth information can effectively separate the foreground objects from the background, and it can be better for the neighbouring objects that are occluded from each other. For data, it also needs to be taken into account, e.g. infrared-RGB multimodal datasets have been constructed\cite{sun2022drone}, but datasets with depth information are still lacking. In addition, it is also necessary to build a wider UAV image object detection dataset, but the efficiency of manual labeling is too low, and a more efficient labeling process needs to be built. In terms of application, these studies are beneficial to the development of other fields, such as UAV control and autonomous driving. These techniques, data and application exploration are worth exploring in the future. However, these are beyond the scope of this paper.

{\small
\bibliographystyle{unsrt}
\bibliography{egbib}
}

\vfill

\end{document}